\documentclass[preprint]{elsarticle}
\usepackage{hyperref}
\usepackage{amsmath}

\usepackage[utf8]{inputenc}
\usepackage[T1]{fontenc}
\usepackage{lmodern}
\usepackage{units}
\usepackage{setspace}
\usepackage{amsfonts}
\usepackage{amssymb}
\usepackage{bm}
\usepackage[dvipsnames]{xcolor}
\usepackage{tabularx}
\usepackage{lipsum}
\usepackage{dsfont}

\newcommand{\round}[1]{\ensuremath{\lfloor#1\rceil}}

\usepackage[]{natbib}
\biboptions{numbers,sort&compress}

\usepackage{algorithm}
\usepackage{algpseudocode}

\usepackage{graphicx}
\usepackage{lineno}

\usepackage{csvsimple}
\usepackage{booktabs}

\usepackage{tikz}
\usepackage{pgfplots}
\usepackage{pgfplotstable}
\usetikzlibrary{arrows,arrows.meta,shapes,positioning,3d}

\usepackage[normalem]{ulem}

\usepackage{fullpage}

\begin{document}

\begin{frontmatter}

\title{Reliability Analysis of Complex Systems using Subset Simulations with Hamiltonian Neural Networks}


\author[]{Denny Thaler\corref{cor}$^{1,2}$}
\ead{thaler@iam.rwth-aachen.de}
\author{Somayajulu LN Dhulipala$^3$}
\author{Franz Bamer$^1$}
\author{Bernd Markert$^1$}
\author{Michael D. Shields$^2$}

\address{Institute of General Mechanics, RWTH Aachen University, 52062 Aachen, Germany}
\address{Department of Civil and Systems Engineering, Johns Hopkins University, 3400 N. Charles St., Baltimore, MD 21218, USA}
\address{Computational Mechanics and Materials, Idaho National Laboratory, Idaho Falls, ID 83402, USA}

\cortext[cor]{Corresponding author.}

\begin{abstract}

We present a new Subset Simulation approach using Hamiltonian neural network-based  Monte Carlo sampling for reliability analysis. The proposed strategy combines the superior sampling of the Hamiltonian Monte Carlo method with computationally efficient gradient evaluations using Hamiltonian neural networks. This combination is especially advantageous because the neural network architecture conserves the Hamiltonian, which defines the acceptance criteria of the Hamiltonian Monte Carlo sampler. Hence, this strategy achieves high acceptance rates at low computational cost. Our approach estimates small failure probabilities using Subset Simulations. However, in low-probability sample regions, the gradient evaluation is particularly challenging. The remarkable accuracy of the proposed strategy is demonstrated on different reliability problems, and its efficiency is compared to the traditional Hamiltonian Monte Carlo method. We note that this approach can reach its limitations for gradient estimations in low-probability regions of complex and high-dimensional distributions. Thus, we propose techniques to improve gradient prediction in these particular situations and enable accurate estimations of the probability of failure. The highlight of this study is the reliability analysis of a system whose parameter distributions must be inferred with Bayesian inference problems. In such a case, the Hamiltonian Monte Carlo method requires a full model evaluation for each gradient evaluation and, therefore, comes at a very high cost. However, using Hamiltonian neural networks in this framework replaces the expensive model evaluation, resulting in tremendous improvements in computational efficiency.

\end{abstract}


\end{frontmatter}

\section{Introduction}


Engineers are responsible for designing reliable structures considering the cost and consequences of structural damage/failure. These structures require different thresholds for the probability of failure. Irreplaceable parts and those whose failure will have dire consequences, i.e., components in space structures or nuclear power plants, need to be constructed such that failure is extremely unlikely. Since structures generally behave nonlinearly before they fail and distributions for uncertain variables may deviate significantly from Gaussian, evaluating the failure probability is usually not straightforward. That is, the failure probability cannot be solved analytically such that numerical and/or statistical reliability methods are required \cite{Dilevsen_2007book}.

The Monte Carlo method is the benchmark statistical method since it provides an unbiased estimate of the failure probability. However, it requires a huge number of model evaluations to provide a reasonable estimate, especially if the probability of failure is very low.  Therefore, it is necessary to significantly speed up this method. Methods aimed at significantly decreasing computational cost generally fall into two major categories: (i) methods that speed up each sample evaluation and (ii) methods that improve sampling efficiency and reduce the number of necessary model evaluations. To speed up model evaluations, it is common to employ model order reduction strategies, e.g.,~\cite{Kerschen2005_POD_overview, Chinesta_2011_POD_review, Bamer2012_POD, Jensen_2016_MOR_UQ, Bamer2017_EffRespIdent,  Skandalos_2022_MulitFide_surrogate, Bamer2017_POD_MC} or construct a machine-learned surrogate model that approximates the model response, e.g.,~\cite{bichon2008efficient,echard2011ak, Thaler2021_Pamm, 
 Sundar2016,schobi2017rare, Thaler2020_Pamm, Li_2020_SurrogateUQ, Bamer2021_Frontiers}. Other approaches combine these strategies in a multi-fidelity framework wherein data are fused from reduced order models and high-fidelity simulations~\cite{peherstorfer2016multifidelity,kramer2019multifidelity,pham2022ensemble, Dhulipala2022_Multifidelity}. 
To improve sampling efficiency, numerous strategies have been developed. Established methods include importance sampling~\cite{au1999new}, Latin Hypercube sampling~\cite{olsson2003latin}, and Subset Simulation~\cite{au2001estimation}, among others.


In this paper, we will focus on a specific aspect of the second strategy, namely efficient sampling. 
Our approach is rooted in the widely-used Subset Simulation method, which -- although not perfect~\cite{breitung2019geometry} -- has become a standard approach for Monte Carlo-based reliability analysis~\cite{au2001estimation, Au2003, Au2007}. Subset Simulation requires Markov Chain Monte Carlo (MCMC) methods to sample from the conditional distributions at each level of the method. Initially, this was done with a modification to the random walk-based Metropolis-Hastings sampler~\cite{Metropolis1953, Hastings1970} to perform component-wise steps~\cite{au2001estimation}. Since that time, numerous schemes have been proposed that use enhanced MCMC methods to improve efficiency, including repeated generation of pre-candidate states~\cite{santoso2011modified}, delayed rejection~\cite{miao2011modified}, and conditional sampling~\cite{papaioannou2015mcmc}. Zuev et al.\ also proposed a Bayesian postprocessor for Subset Simulation to further estimate the probability density function to quantify the uncertainty of the failure probability~\cite{zuev2012bayesian}. 

The aforementioned methods generally consider Subset Simulation formulated in the standard Gaussian space. However, performing Subset Simulation in the original physical space is often desirable because an isoprobabilistic transformation to standard Gaussian is not easily obtained. In the original space, however, Subset Simulation can be complicated by strong nonlinear dependence among variables coupled with strong non-Gaussianity and degeneracy of high-dimensional distributions. Two recent approaches have specifically aimed to address this issue. Shields et al.~\cite{shields2021subset} applied the random walk-based affine invariant ensemble ``Stretch'' sampler in Subset Simulation for problems where sampling the conditional distributions may be difficult.  Meanwhile, Wang et al.~\cite{wang2019hamiltonian} proposed algorithms that leverage Hamiltonian Monte Carlo (HMC) to generate samples by simulating the evolution of time-reversible Hamiltonian dynamics using a symplectic numerical integrator. This work was further extended to use Riemannian manifold HMC in~\cite{chen2022riemannian}. 


The major disadvantage of HMC compared to Metropolis-Hastings and other random walk methods is the computational cost. The proposal of a new state in HMC requires simulating the evolution of time-reversible Hamiltonian dynamics numerically using symplectic integration. Therein, evaluating gradients of the Hamiltonian causes the greatest computational effort. To improve the efficiency, Strathmann et al.~\cite{strathmann2015gradient} proposed a gradient-free approach based on exponential kernels. Li et al.~\cite{Li2019} proposed to use neural networks to approximate the numerical gradient in HMC, while 
Levy et al.~\cite{levy2018generalizing} proposed to learn a neural network operator that serves as an efficient kernel for the HMC method. These neural network methods, however, were not constrained by the Hamiltonian dynamics. Greydanus et al.~\cite{Greydanus2019} proposed the Hamiltonian neural network (HNN), a physics-informed neural network that conserves energy over long trajectories.
Recently, Dhulipala et al.~\cite{Dhulipala2022} proposed to use HNNs for efficient HMC sampling for Bayesian inference and its integration with an advanced version of HMC called the No-U-Turn Sampler (NUTS). Dhulipala et al.~\cite{Dhulipala2022} also proposed a modified version of HNNs called latent HNNs. This acceleration of HMC with HNNs has been further verified in~\cite{Thaler2022_PammHNNMC}.

In this paper, we propose to integrate pre-trained HNNs with HMC, termed Hamiltonian Neural Network Monte Carlo (H\textsubscript{NN}MC), to accelerate Subset Simulations. The proposed method provides the same robust probability of failure estimates using the standard HMC methods, earlier proposed by Wang et al.~\cite{wang2019hamiltonian}, while reducing the cost of the Hamiltonian integration considerably. We achieve greater than 20 times speedup in the propagation of the Hamiltonian trajectories, which makes conditional sampling in Subset Simulation far more efficient than the standard HMC. Furthermore, we provide strategies to overcome the limitations of pretraining the HNNs that make extensions of the proposed H\textsubscript{NN}MC algorithm possible for even more challenging problems such as those where Bayesian inference and reliability analysis are conducted together. The final example of this paper explores this special case in which model evaluations are required at every HMC step. However, the proposed H\textsubscript{NN}MC predicts the gradient directly, such that model evaluations are not needed for Hamiltonian integration. In fact, model evaluation is only required once to accept or reject the conditional state, as in the conventional standard Subset Simulation, resulting in a significant speed-up.

Section~\ref{sec:SS_RA} briefly introduces Subset Simulations before the H\textsubscript{NN}MC method is presented in Section~\ref{sec:HNNMC}. Using H\textsubscript{NN}MC within Subset Simulations is proposed in Section~\ref{sec:HNN_SuS} followed by a proof of concept in Section~\ref{sec:PoC}. Section~\ref{sec:SS_BI} demonstrates the strategy on systems with uncertain parameters. Finally, Section~\ref{sec:Conclusion} concludes the results of the proposed strategy. 


\section{Subset Simulation for Reliability Analysis}\label{sec:SS_RA}


To determine whether structural failure occurs, one must define a limit state based on engineering decisions. 
The limit state function $g(\mathbf{x})$ is defined as a function of the random vector $\mathbf{x}$ describing uncertainties in the system and its inputs such that $g(\mathbf{x})\le 0$  
corresponds to failure of the system. The probability of failure is then determined by the multi-dimensional integral of the probability density function $f_{\!X}(\mathbf{x})$ over the failure domain~\cite{Bucher2009_book}:
\begin{eqnarray} \label{eq:probability_of_failure}
    P_{\!F} = P(F) = \underset{g(\mathbf{x})\leq 0}{\int \cdots \int} f_{\!X}(\mathbf{x}) \; \text{d}\mathbf{x} \; \text{,}
\end{eqnarray}
where $F$ is the failure region corresponding to the event $g(\mathbf{x})\leq 0$. The equation can be reformulated using an indicator function $\mathds{I}_F(\mathbf{x})$, for which  $\mathds{I}_F(\mathbf{x})=1$ if $\mathbf{x} \in F$ and $\mathds{I}_F(\mathbf{x})=0$ otherwise, as follows:
\begin{equation} \label{eq:MC_probability}
    P_F = \int_{-\infty}^\infty  \dots \int_{-\infty}^\infty \mathds{I}_F(\mathbf{x}) f_{\!X}(\mathbf{x}) \; \text{d}\mathbf{x} \;\text{.} \;
\end{equation}

Using standard Monte Carlo methods, the probability of failure can be estimated by the expectation of the indicator function, written as:
\begin{equation} \label{eq:MC_expectation}
    P_F = E[\mathds{I}_F(\mathbf{x})] \approx \frac{1}{m}\sum_{k=1}^{m} \mathds{I}_F(\mathbf{x}^{(k)}) \; \text{.}
\end{equation}
where $\mathbf{x}^{(k)}$ are $m$ independent samples drawn from $f_{\!X}(\mathbf{x})$. 
In general, this method gives accurate and robust probability of failure estimates, but it converges slowly -- requiring large numbers of samples in which $m\propto\dfrac{1}{P_F}$. For small $P_F$, this is clearly problematic.

Subset Simulation, proposed by Au and Beck~\cite{au2001estimation}, aims to reduce the number of samples by defining $P_{\!F}$ through a
nested sequence of failure regions $F_1 \supset F_2 \supset \dots \supset F_n = F$ so that $F = \cap ^k_{i=1} F_i$, $k =1, \dots, n$.  The probability of failure is then evaluated using conditional probabilities as follows:
\begin{align}
        P(F) = 
        P(F_1)  \prod_{i=1}^{n-1} P(F_{i+1} \mid F_i)\; \text{.}
\end{align}
The conditional probabilities are set to be larger values, typically around 0.1, and are estimated using Monte Carlo simulation by drawing samples from the conditional distributions at each level. These samples are drawn using various MCMC algorithms to be constrained within the conditional subsets. As previously mentioned, a great deal of research has focused on using different MCMC algorithms for this task. Of particular interest in this work is the use of HMC, which we discuss next.




\section{Hamiltonian Neural Networks in Hamiltonian Markov Chain}\label{sec:HNNMC}

In this section, we first provide a brief overview of the Hamiltonian Monte Carlo method. Next, we introduce Hamiltonian Neural Networks (HNNs), a type of physics-informed neural network specifically designed to learn Hamiltonian dynamics. We then show how HNNs can be integrated into HMC to accelerate the sampling process.

\subsection{Hamiltonian Monte Carlo}

Hamiltonian Monte Carlo makes the analogy between sampling and a set of particles moving through a probability distribution, $\pi(\mathbf{x})$, with no external forces (i.e., only conservative forces). Under these conditions, the moving particles are governed by the following principles of Hamiltonian mechanics, wherein the state of the particles can be fully described by the position vector $\mathbf{q}$ and the momentum vector $\mathbf{p}$. Since the system is conservative, its energy (Hamiltonian) is constant and composed of the kinetic and potential energy, $K$ and $U$, which depend only on the position and momentum. The Hamiltonian is written as follows:
\begin{equation}
    H(\mathbf{q},\mathbf{p})= U(\mathbf{q}) + K(\mathbf{p})\; \text{.}
\end{equation}
The state of the system evolves with time using Hamilton's equations relating the position $\mathbf{q}$ and the momentum $\mathbf{p}$ as:
\begin{align}
    \frac{d \mathbf{q}}{dt} &=\frac{\partial H}{\partial \mathbf{p}} \nonumber \; \text{,} \\
    \frac{d \mathbf{p}}{dt} &= -\frac{\partial H}{\partial \mathbf{q}}\; \text{.}
    \label{eqn:Hamltons_Eqns}
\end{align}
Solving this system of equations in time results in a constant energy (Hamiltonian) trajectory in the $(\mathbf{p},\mathbf{q})$ phase space. Importantly for HMC, these equations describe a system that is reversible and symplectic (i.e., conserves energy and preserves volume). The importance of these properties for HMC has been detailed in other works~\cite{Neal2012, wang2019hamiltonian} and will not be discussed further here.

HMC leverages Hamiltonian dynamics by treating the sample $\mathbf{x}$ drawn from $\pi(\mathbf{x})$ as a set of particles having positions $\mathbf{q}$ in the Hamiltonian trajectory (i.e., $\mathbf{x}\equiv \mathbf{q}$). We then define the joint probability density between the position $\mathbf{q}$ and momentum $\mathbf{p}$ to take the following form:
\begin{eqnarray}
    \pi(\mathbf{p},\mathbf{q}) \propto e^{-H(\mathbf{p},\mathbf{q})} = e^{-U(\mathbf{q})}e^{-K(\mathbf{p})}\; \text{.}
\end{eqnarray}
We then define the potential energy by: 
\begin{equation}
    U(\mathbf{q})= -\log{\pi(\mathbf{q})}\; \text{,}
\end{equation}
and the kinetic energy by:
\begin{equation}
    K(\mathbf{p}) = \frac{1}{2}\mathbf{p} \mathbf{M}^{-1}\mathbf{p}\; \text{,}
\end{equation}
which results in the position following the target distribution $\pi(\mathbf{q})$ and the momentum independently following a Gaussian distribution with covariance $\mathbf{M}$.  Selecting $\mathbf{M}$ to be a non-diagonal, positive semi-definite covariance matrix results in the Riemannian Manifold HMC, which has been demonstrated to improve the acceptance rate for non-Gaussian distributions but requires determining the best $\mathbf{M}$ matrix for a given problem~\cite{Betancourt2018, chen2022riemannian}. However, in this work, as is often done, we select $\mathbf{M}=\alpha\mathbf{I}$ where $\mathbf{I}$ is the identity matrix, and $\alpha$ is a scalar value, which corresponds to the momenta being independent Gaussian random variables.

To perform HMC, we then consider the current state of the Markov chain as the position $\mathbf{q}$ and assign a Gaussian random momentum $\mathbf{p}$, which completely defines a state in phase space such that Eq.~\eqref{eqn:Hamltons_Eqns} defines a reversible trajectory of constant energy (Hamiltonian). Hamilton's equations (Eq.\ \eqref{eqn:Hamltons_Eqns}) are then solved numerically using a symplectic integrator. Here, we use the synchronized leapfrog scheme, which first updates the positions written as:
\begin{equation}
    \mathbf{q}(t+\Delta t) = \mathbf{q}(t) + \frac{\Delta t}{m_i}~\mathbf{p}(t) + \frac{\Delta t^2}{2m_i}~\dot{\mathbf{p}}(t) = \mathbf{q}(t) + \frac{\Delta t}{m_i}~\mathbf{p}(t) - \frac{\Delta t^2}{2m_i}~\frac{\partial H}{\partial \mathbf{q}(t)} \; \text{,}
    \label{eqn:leapfrog1}
\end{equation}
where $\Delta t$ is the integration time step and $\dot{\mathbf{p}}(t)=-\frac{\partial H}{\partial \mathbf{q}(t)}$. Then, the momenta are updated by:
\begin{equation}
    \mathbf{p}(t+\Delta t) = \mathbf{p}(t) + \frac{\Delta t}{2}~\bigg(\dot{\mathbf{p}}(t) + \dot{\mathbf{p}}(t+\Delta t)\bigg) = \mathbf{p}(t) - \frac{\Delta t}{2}~\bigg(\frac{\partial H}{\partial \mathbf{q}(t)} + \frac{\partial H}{\partial \mathbf{q}(t+\Delta t)}\bigg) \; \text{,}
    \label{eqn:leapfrog2}
\end{equation}
where again $\dot{\mathbf{p}}(t)=-\frac{\partial H}{\partial \mathbf{q}(t)}$.


A new state $(\mathbf{q^*},\mathbf{p^*})$ is determined by integrating forward $L$ time steps for a total time of $t_f=L\Delta t$. Because the integration is performed numerically, there is some error that causes a deviation in the Hamiltonian (i.e., \ $H$ is not strictly conserved). Therefore, a Metropolis-Hastings type acceptance-rejection criterion is introduced that accepts the new state with probability 
\begin{equation}
\alpha = \min \left[ 1, \exp( H (\mathbf{q}, \mathbf{p}) - H(\mathbf{q^*}, \mathbf{p^*})) \right] \; \text{.}
\end{equation}

The HMC process is highly sensitive to both $L$ and $\Delta t$ such that large $\Delta t$ can cause large integration errors and thus high rejection rates. Meanwhile, small $L$ can result in undesirable random-walk-type behavior, while large $L$ can result in unnecessarily high computation costs. To avoid the need to manually tune the parameters, one may apply methods such as the No-U-Turns Sampler (NUTS)~\cite{hoffman2014no}. However, in the context of Subset Simulation, Wang et al.~\cite{wang2019hamiltonian} proposed methods for setting $t_f$, which will be discussed in Section \ref{sec:HNN_SuS}. Here, we apply strategies similar to NUTS to generate the samples from the original distribution and rely on methods of Wang et al.\ for sampling from conditional distributions at subsequent levels.  

For completeness, the Hamiltonian Monte Carlo method is detailed in Algorithm~\ref{alg:HMC}.

\begin{algorithm}[!ht]
\caption{Hamiltonian Monte Carlo}\label{alg:HMC}
\begin{algorithmic}
\State Initial state: $\textbf{q}$, Hamiltonian: $H$, Leapfrog steps: $L$, Step size: $\Delta t$
\State $\textbf{p}_0 \gets \mathcal{N}(0,1)$
\State $\textbf{q}_0 \gets q$
\State $\textbf{p}_{\frac{1}{2}} \gets \textbf{p}_0 - \frac{\Delta t}{2} \frac{\partial H}{\partial \textbf{q}}$
\For{$ 1 \leq n < L$} 
    \State $\textbf{q}_{n} \gets \textbf{q}_{n-1} + \Delta t p_n$
    \State $\textbf{p}_{n+\frac{1}{2}} \gets \textbf{p}_{n-\frac{1}{2}} - \Delta t \frac{\partial H}{\partial \textbf{q}}$
\EndFor
\State $\textbf{p}_{L} \gets p_{L-1} - \frac{\Delta t}{2} \frac{\partial H}{\partial q}$
\State $\textbf{q}* \gets \textbf{q}_L$
\State $\textbf{p}* \gets -\textbf{p}_L$
\State $\alpha = \min{\left[ 1, \exp{\left(-H(\textbf{q}*,\textbf{p}*)+H(\textbf{q}_0,\textbf{p}_0)\right)}\right]}$
\If{$\alpha \geq \mathcal{U}(0,1)$ }
   \State \Return $(\textbf{q}*,\textbf{p}*)$
\Else
    \State \Return $(\textbf{q}_0,\textbf{p}_0)$
\EndIf

\end{algorithmic}
\end{algorithm}

\subsection{Hamiltonian Neural Networks}

\begin{figure}[!ht]
    \centering
    \includegraphics{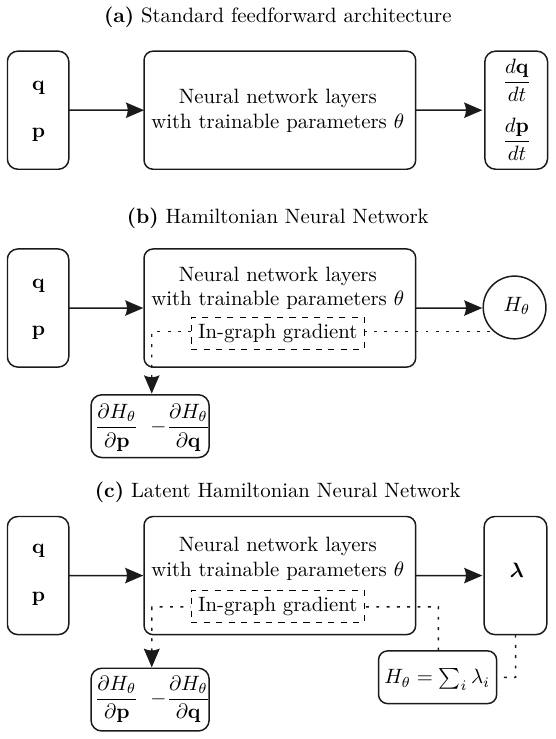}
    \caption{Comparison of the neural network architectures: (a) Scheme of a standard feedforward neural network; (b)  Hamiltonian Neural Network architecture with an in-graph gradient, cf.~\cite{Greydanus2019}; (c) Scheme of latent Hamiltonian Neural Networks, cf. \cite{Dhulipala2022}.}
    \label{fig:HNN_scheme}
\end{figure}

HMC requires the computation of the gradient of the system $\left( \frac{\partial H}{\partial \mathbf{q}} \right)$ at each time step, which can be computationally expensive, and each proposal of HMC requires many time steps. To reduce this cost, gradient evaluation can be accelerated using a surrogate model. Li et al.~\cite{Li2019} showed that standard feedforward neural networks, such as that shown in Fig.~\ref{fig:HNN_scheme}(a), result in a high speed-up. However, recent enhancements in physics-informed neural networks that conserve energy promise high accuracy in long-run time integration of Hamiltonian systems. One example of particular interest is the Hamiltonian Neural Network (HNN)~\cite{Greydanus2019}. The idea of HNNs is to learn the Hamiltonian $H(\mathbf{q},\mathbf{p})$ such that energy is conserved -- that is, Hamilton's equations are satisfied. This is achieved by constructing a neural network that learns the parameters $\theta$ by solving
 \begin{equation}
\underset{\theta}{\operatorname{argmin}}\left\|\frac{\mathrm{d} \mathbf{q}}{\mathrm{dt}}-\frac{\partial H_\theta}{\partial \mathbf{p}}\right\|^2+\left\|\frac{\mathrm{d} \mathbf{p}}{\mathrm{dt}}+\frac{\partial H_\theta}{\partial \mathbf{q}}\right\|^2,
\label{eqn:HNN_loss}
\end{equation}
where the gradients $\frac{\partial H_\theta}{\partial \mathbf{p}}$ and $\frac{\partial H_\theta}{\partial \mathbf{q}}$ are obtained directly from the neural network through automatic differentiation as illustrated in Fig.~~\ref{fig:HNN_scheme}(b). Recently, a variation of HNNs, termed Latent HNNs (L-HNNs), was proposed in which the network output is a set of $d$ latent variables $\boldsymbol{\lambda}=\{\lambda_1, \lambda_2,\dots, \lambda_d\}$ where $d$ is the number of random variables such that 
\begin{equation}
    H_\theta = \sum_{i=1}^d \lambda_i \; \text{,}
    \label{eqn:L-HNN}
\end{equation}
which has been shown to improve the expressivity of the network \cite{Dhulipala2022}. The L-HNN structure is shown in Fig.~\ref{fig:HNN_scheme}(c). Next, we will briefly discuss the nuances of training these different neural networks and their implications for HMC.

The standard feedforward neural network approach takes as input the position and momentum $(\mathbf{q},\mathbf{p})$ and returns the gradients $(\frac{\mathrm{d}\mathbf{q}}{\mathrm{d}t}, \frac{\mathrm{d}\mathbf{p}}{\mathrm{d}t})$ directly. Hence, training data consists of trajectories of Hamilton's equations, where each time step provides a training point. In HNNs, the training data are the same, but the network is constructed to learn a function $H_\theta$ intended to approximate the Hamiltonian. By establishing the loss function in Eq.~\eqref{eqn:HNN_loss}, the network learns $H_\theta$ such that energy is conserved by comparing the gradients of $\frac{\partial H_\theta}{\partial \mathbf{p}}$ and $\frac{\partial H_\theta}{\partial \mathbf{q}}$ with the gradients $(\frac{\mathrm{d}\mathbf{q}}{\mathrm{d}t}, \frac{\mathrm{d}\mathbf{p}}{\mathrm{d}t})$ in the training data. Finally, the L-HNN takes advantage of the fact that the gradient is a linear operation and instead predicts a set of $d$ latent variables from which $H_\theta$ can be computed by Eq.~\eqref{eqn:L-HNN}, thus improving expressivity, particularly for high-dimensional problems. We therefore notice that, although the training data are identical, the HNNs and L-HNNs will have far superior performance -- especially for long trajectories -- because they are designed to conserve energy. 

\subsection{The H\textsubscript{NN}MC Algorithm}\label{section:HNNMC}

Integration of HNNs into HMC is straightforward but can have significant computational benefits. This method, which we refer to as H\textsubscript{NN}MC simply replaces the standard numerical gradient evaluation in the leapfrog integration (Eqs.~\eqref{eqn:leapfrog1} and \eqref{eqn:leapfrog2}) with HNN approximated gradients $\frac{\partial H_\theta}{\partial \mathbf{p}}$ and $\frac{\partial H_\theta}{\partial \mathbf{q}}$. This way, the Hamiltonian trajectories necessary for generating the next state of the Markov chain can be computed without the expensive numerical gradients that would typically be required when performing the conventional approach. The resulting algorithm is provided in Algorithm~\ref{alg:HNNMC}. 

Of course, the HNN requires sufficient training data, and in its application to HMC, this training data is generated \textit{a priori}. Consequently, the cost of the H\textsubscript{NN}MC is incurred up-front, and the subsequent chain propagation comes at little cost. Generally, this up-front cost is far less than the cost of numerical integration and it becomes increasingly efficient as the number of time steps and the number of samples increases.

\begin{algorithm}[!h]
\caption{Hamiltonian Neural Network Monte Carlo}\label{alg:HNNMC}
\begin{algorithmic}
\State Initial state: $\textbf{q}$, Hamiltonian: $H$, Leapfrog steps: $L$, Step size: $\Delta t$, Hamiltonian Neural Network prediction: $\operatorname{HNN(\dot)}$, Neural network input: $\mathbf{x}=\mathbf{q}\oplus\mathbf{p}$ 
\State $\textbf{p}_0 \gets \mathcal{N}(0,1)$
\State $\textbf{q}_0 \gets \textbf{q}$
\State $\textbf{p}_{\frac{1}{2}} \gets \textbf{p}_0 - \frac{\Delta t}{2} \operatorname{HNN}(\mathbf{x})$
\For{$ 1 \leq n < L$} 
    \State $\textbf{q}_{n} \gets \textbf{q}_{n-1} + \Delta t \textbf{p}_n$
    \State $\textbf{p}_{n+\frac{1}{2}} \gets \textbf{p}_{n-\frac{1}{2}} - \Delta t \operatorname{HNN}(\mathbf{x})$
\EndFor
\State $\textbf{p}_{L} \gets \textbf{p}_{L-1} - \frac{\Delta t}{2} \operatorname{HNN}(\mathbf{x})$
\State $\textbf{q}* \gets \textbf{q}_L$
\State $\textbf{p}* \gets -\textbf{p}_L$
\State $\alpha = \min{\left[ 1, \exp{\left(-H(\textbf{q}*,\textbf{p}*)+H(\textbf{q}_0,\textbf{p}_0)\right)}\right]}$
\If{$\alpha \geq \mathcal{U}(0,1)$ }
   \State \Return $(\textbf{q}*,\textbf{p}*)$
\Else
    \State \Return $(\textbf{q}_0,\textbf{p}_0)$
\EndIf

\end{algorithmic}
\end{algorithm}

\subsection{H\textsubscript{NN}MC Illustration}\label{section:HNNMC_illustration}

In this section, we briefly demonstrate the use of HNNs in HMC. To achieve good results, it is crucial to accurately reproduce the Hamiltonian system. First, we show that the HNN accurately conserves the Hamiltonian for a 1D bimodal distribution. We then show that the HNN gradients produce accurate Hamiltonian trajectories for a 2D distribution with correlations. 

For the first demonstrative example, we choose a bimodal Gaussian with $\boldsymbol{\mu} = [0, 3] $ and $\boldsymbol{\sigma} = [1, 1]$ as the target distribution, where the second peak has a higher weight, having a pdf given by:
\begin{equation}
    p(q) = \frac{\sigma}{\sqrt{2\pi}} \left( \frac{1}{4} \exp \left( -0.5  \left(\frac{q-\mu_1}{\sigma}\right)^2\right) + \frac{3}{4} \exp\left( -0.5  \left(\frac{q-\mu_2}{\sigma} \right)^2 \right) \right) \;\text{.}
\end{equation}

\begin{figure*}[!ht]
    \centering
    \includegraphics[width=1.\textwidth]{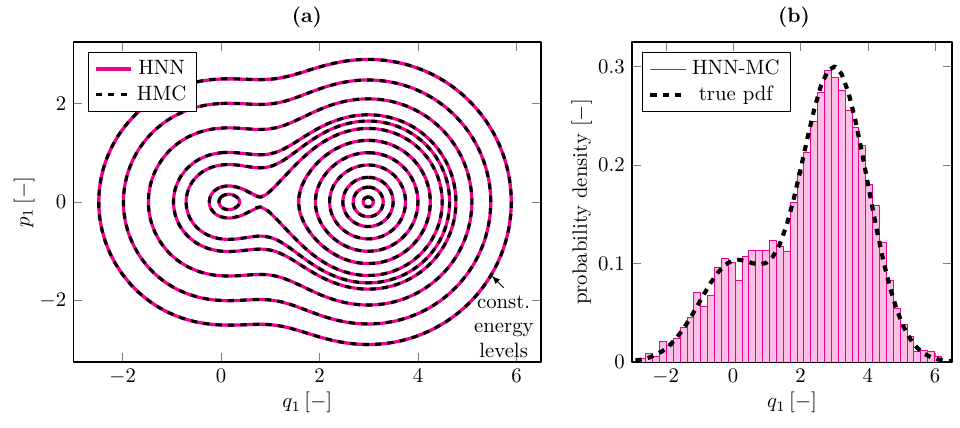}
    \caption{Application of  H\textsubscript{NN}MC for a bimodal Gaussian mixture distribution. (a) Trajectories in phase space of constant Hamiltonian. The trajectories are simulated using the gradients of the Hamiltonian (HMC) or the predicted values (HNN); (b) The distribution of $5000$ samples using the H\textsubscript{NN}MC compared with the true probability density function.}
    \label{fig:1D_bimod_traj}
\end{figure*}

The training data are generated by simulating relatively long training trajectories from a few initial samples. For this distribution, $40$ trajectories are simulated with a trajectory length of $20$ and a step size of $0.05$. The HNN consists of two hidden layers with ten neurons in each layer. This relatively simple architecture can achieve high training and test accuracy. For this one-dimensional problem, the accurate prediction of the HNN can be seen in Figure~\ref{fig:1D_bimod_traj}(a) by comparing the simulated trajectories with those using traditional gradient calculation in phase space.  
The Hamiltonian is well preserved using the HNN, as the constant energy levels are simulated accurately. 
Figure \ref{fig:1D_bimod_traj}(b) shows the distribution estimated from $5\,000$ samples generated using the H\textsubscript{NN}MC matches the true probability density with high accuracy.

For the second illustrative example, we apply H\textsubscript{NN}MC for a two-dimensional correlated Gaussian distribution that has the mean $\boldsymbol{\mu} = [1, -1]^{T} $ and the covariance matrix $\boldsymbol{\Sigma} =  \begin{bmatrix}
1. & 0.9 \\
0.9 & 1.
\end{bmatrix}$, with pdf given by:
\begin{equation}
    p(\mathbf{q}) = \frac{\exp \left(
    -0.5  
    \left(
    \mathbf{q}-\boldsymbol{\mu} \right)^T \mathbf{\Sigma}^{-1} \left(
    \mathbf{q}-\boldsymbol{\mu} \right)
    \right)}
    {\sqrt{(2\pi)^2 \left\lVert \mathbf{\Sigma} \right\rVert }}  \;\text{.}
\end{equation}

\begin{figure*}[!ht]
    \centering
    \includegraphics[width = \textwidth]{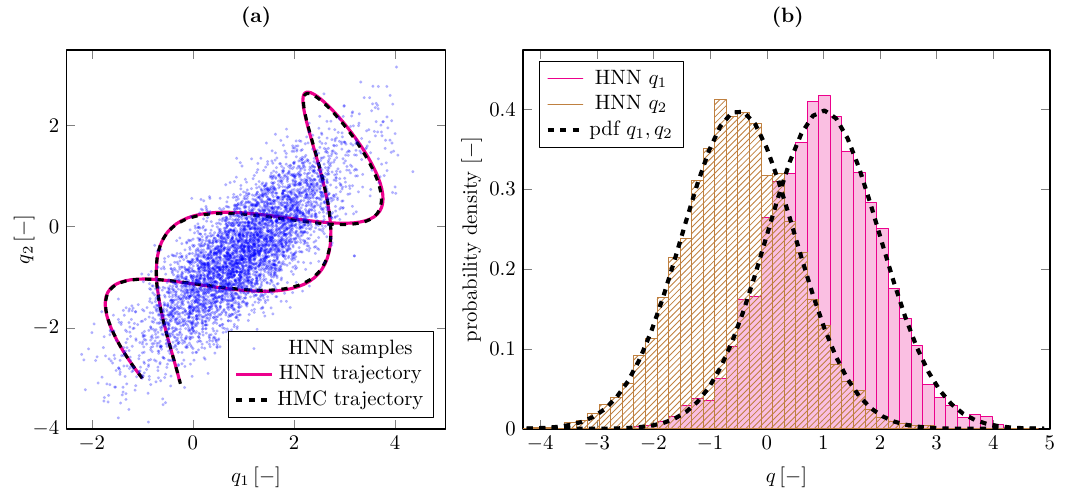}
    \caption{Two-dimensional correlated Gaussian distribution: (a) $5000$ generated samples along with a long trajectory that moves through the distribution using traditional and HNN predicted gradients. (b) Probability density of the variables using $5000$ samples compared to the true probability density function.}
    \label{fig:2D_corr_gauss}
\end{figure*}

The generated samples of the correlated distribution using H\textsubscript{NN}MC are shown in Figure \ref{fig:2D_corr_gauss}(a). To compare the gradient evaluations, one trajectory is simulated using the standard Hamiltonian gradients as well as the predicted gradients using HNNs. To show the accuracy of long-run trajectories, the length is chosen larger than the required trajectories during sampling. 
We observe that the HNN trajectory follows the original one and conclude that the gradient prediction for this trajectory is accurate.
For both variables, $q_1$ and $q_2$, the probability density functions are shown in Figure~\ref{fig:2D_corr_gauss}(b). Both match the true probability density function of the Gaussian shifted to the respective mean.

\section{H\textsubscript{NN}MC for Subset Simulation}
\label{sec:HNN_SuS}

In this section, we propose a methodology to integrate H\textsubscript{NN}MC into Subset Simulation. The framework closely follows the method proposed by Wang et al.~\cite{wang2019hamiltonian} for the use of HMC in Subset Simulation, so we begin by discussing this approach.

\subsection{HMC for Subset Simulation}

Two primary considerations must be addressed when applying HMC for conditional sampling in Subset Simulation. The first is the fundamental question of sampling from the \textit{conditional} distribution. To account for the truncation of the distribution that occurs when defining each conditional level, Wang et al.\ point out that integrating the conditional distribution into the Hamiltonian as follows:
\begin{equation}
\begin{aligned}
    H(\mathbf{q},\mathbf{p}) = U(\mathbf{q}) + K(\mathbf{p}) & = -\log(\pi(\mathbf{q}|F_j)) + \frac{1}{2}\mathbf{p} \mathbf{M}^{-1}\mathbf{p}\\
    & =-\log(\pi(\mathbf{q})) + \frac{1}{2}\mathbf{p} \mathbf{M}^{-1}\mathbf{p} - \log(\mathbb{I}_F(\mathbf{q})) + \text{const.}
\end{aligned}
\end{equation}
creates a \textit{potential barrier}. That is, when $\mathbb{I}_F(\mathbf{q})=0$, the Hamiltonian has infinite potential energy. 

To account for this energy barrier, they proposed two algorithms. The first approach effectively ignores the barrier and applies a \textit{post hoc} acceptence-rejection step. This scheme, referred to as Rejection Sampling HMC, simply computes the Hamiltonian trajectory to generate the next state $\mathbf{q^*}$ and then rejects this state if $\mathbb{I}_F(\mathbf{q^*})=0$; that is if the new state does not lie in the current conditional level $F_j$.

The second, more complicated approach uses a novel algorithm to approximate the time at which the trajectory crosses the limit surface (potential barrier), calculates the momentum at this ``hitting time,'' and bounces the trajectory off the potential barrier, after which the trajectory continues, and a new state $\mathbf{q^*}$ is proposed. This state is then again rejected if  $\mathbb{I}_F(\mathbf{q^*})=0$. This method, referred to as the Barrier Bouncing HMC, will not be discussed in detail here. In our applications, we apply a Rejection Sampling-based approach for simplicity, although there is no reason that the barrier-bouncing approach could not be applied in our setting. 

The second issue in applying HMC for Subset Simulation is the appropriate selection of the trajectory length in the subsets. As stated previously, one could naively apply the 
HMC using Algorithm~\ref{alg:init-t_f}. However, this may result in relatively long trajectories that wander out of the conditional level, yielding a high rejection rate. On the other hand, selecting very short trajectories yields very high acceptance rates, causing a strong correlation between samples. Wang et al. propose to aim for a relatively constant acceptance rate targeted in the range between $a_{low}=0.3$ and $a_{high}=0.5$. To achieve this target acceptance rate, they proposed the method in Algorithm~\ref{alg:adaptive-t_f}. The initial trajectory length 
 is set based on Algorithm~\ref{alg:init-t_f}.

\begin{algorithm}[!ht]
\caption{Initialize trajectory length and set leapfrog steps}\label{alg:init-t_f}
\begin{algorithmic}
\State Step counter: $j$, Step size: $\Delta t$, Mean period: $\Bar{T}$, Factor for division: $k$ 

\For{$i \gets 1,N$}

\State $(\mathbf{q},\mathbf{p}) \gets (\mathbf{q}^{(i)},\mathbf{p}^{(i)})$
\State $j = 1$
\State $(\mathbf{q_+},\mathbf{p_+}) \gets \text{One leapfrog step} (\mathbf{q},\mathbf{p})$
\State $(\mathbf{q_-},\mathbf{p_-}) \gets \text{One leapfrog step} (\mathbf{q},\mathbf{-p})$

\State
\While{$\mathbf{p}_+^{T}(\mathbf{q}_+-\mathbf{q}_-)$ or $\mathbf{p}_-^{T}(\mathbf{q}_--\mathbf{q}_+)$ } 
\State $j +=1$
\State $(\mathbf{q_+},\mathbf{p_+}) \gets \text{One leapfrog step} (\mathbf{q_+},\mathbf{p_+})$
\State $(\mathbf{q_-},\mathbf{p_-}) \gets \text{One leapfrog step} (\mathbf{q_-},\mathbf{p_-})$

\EndWhile

\State ${T_i} \gets 2j \Delta t$ 

\EndFor
\State
\State $ \Bar{T} \gets \dfrac{\sum{T_i}}{N}$ 
\State $t_f \gets \Bar{T}/k$
\State $ L \gets \round{\dfrac{t_f}{\Delta t}}$ \Comment{Set leapfrog steps to nearest integer}
\State
\State \Return $[\Bar{T},L]$

\end{algorithmic}
\end{algorithm}

\begin{algorithm}[!ht]
\caption{Adaptive rule for trajectory length}\label{alg:adaptive-t_f}
\begin{algorithmic}
\State \textbf{} Subset number: $j$, Step size: $\Delta t$, Mean period: $\Bar{T}$
\If{$a_{low}< a < a_{high}$ }
\State \Return No update for leapfrog steps
\Else
\If{$a_{low}< a$} 
\State $t_f \gets \dfrac{ \Bar{T} }{2 \pi} \sin^{-1}\left(\sin\left(\dfrac{2 \pi t_f }{\Bar{T}}\right)\right) \exp ((a-a_{low})/2))$
\ElsIf{$a_{high}> a$}
\State $t_f \gets \dfrac{ \Bar{T} }{2 \pi} \sin^{-1}\left(\sin\left(\dfrac{2 \pi t_f }{\Bar{T}}\right)\right) \exp ((a-a_{high})/2))$
\EndIf

\EndIf
\State $ L \gets \round{\dfrac{t_f}{\Delta t}}$ \Comment{Set Leapfrog steps to nearest integer}

\end{algorithmic}
\end{algorithm}

\subsection{Proposed Approach}\label{sec:Proposed_Approach}

The proposed approach to integrate H\textsubscript{NN}MC into Subset Simulation proceeds as follows:

\begin{enumerate}
    \item Pretrain an HNN by running some set of Hamiltonian trajectories. 
    \item Generate an initial set of samples according to the probability distribution $\pi(\mathbf{x})$ using H\textsubscript{NN}MC.
    \item For each conditional level, draw samples using either the Rejection Sampling HMC or the Barrier Bouncing HMC. However, within the time integration for the Hamiltonian trajectories, compute the gradients of the Hamiltonian using the pre-trained HNN.
\end{enumerate}

These conceptually simple steps are implemented using the pieces described above and can be augmented with certain optional enhancements as described next.

\subsubsection{Online Error Monitoring}\label{section:error_monitoring}
 A general limitation of neural networks is that they interpolate well but often extrapolate inaccurately~\cite{Thaler2021_EQE, Thaler2022_Selection}. Hence, circumstances arise where the HNN gradient predictions become inaccurate as a result of insufficient training data. This is particularly true in the tails of the distribution whose states have not previously been observed. For such cases, Dhulipala et al.~\cite{Dhulipala2022} proposed an online error monitoring scheme based on slice sampling, which was originally proposed to monitor integration errors during the standard HMC.  At the current state, a slice variable $u$ is drawn from the Uniform distribution written as:
 \begin{equation}
     u \sim \mathcal{U}(0,\exp({H\left(\textbf{q}_0,\textbf{p}_0)\right)}) \;\text{,}
 \end{equation}
where $H(\textbf{q}_0,\textbf{p}_0)$ is the Hamiltonian at the current state. Defining a threshold $\Delta \varepsilon$, if
\begin{equation}
    H\left(\textbf{q}_n,\textbf{p}_n)\right) + \ln(u) >  \Delta \varepsilon \; \text{,}
\end{equation}
then the algorithm reverts to conventional numerical integration for the current time step. This increases computational cost but improves robustness, particularly for strongly non-Gaussian and high-dimensional distributions where exploration is difficult.

\subsubsection{Neural network updates}\label{section:param_update}

Aside from the error monitoring scheme, the neural network parameters can be updated during the Subset Simulations. For example, whenever a new subset level is reached, a share of the new samples can be used to create new training data using the traditional gradient calculation. Afterward, the data can be used to retrain the neural network. This procedure leads to higher accuracy in the region of interest and, therefore, reduces the number of rejections.  

Another possible approach is to use the online error monitoring scheme and neural network updates together. In such cases, based on the history of online error monitoring, the scheme decides whether a neural network update is necessary, e.g., when an error monitoring threshold has been exceeded several times, neural network retraining may be triggered.

\subsection{Benefits of H\textsubscript{NN}MC for Subset Simulation}

The main benefit of using H\textsubscript{NN}MC is the computational savings compared to standard HMC. Our first examples will show the significant speed-up of the sampling while the limit state function of the chosen problems is rather cheap. In these examples, we demonstrate the performance of H\textsubscript{NN}MC on complex distributions and show that the HNNs are able to estimate gradients for rare events. 

In general, most computational expense comes from model evaluation, i.e., the evaluation of the limit state function. To this end, Wang et al.\ state that ``in practice, the main computational effort in Subset Simulation is usually the evaluation of limit-state functions, and each leapfrog step (except the last step) does not involve limit-state function evaluation, thus the additional cost introduced by using a relatively small $\Delta t$ is often negligible'' \cite{wang2019hamiltonian}. This is true for problems in which the probability density function is well-known a priori, as in the traditional Subset Simulation methods where all computations are performed using standard normal random variables. However, this is not the case in the more general reliability problem in which the probability density is known only implicitly (i.e. through the computational model), known only up to a scale factor, or must be inferred from data. In such cases, for example in inference problems where the system parameters are unknown, the model needs to be evaluated to obtain the gradients required for HMC. Consequently, each leapfrog step becomes extremely expensive. H\textsubscript{NN}MC, on the other hand, is capable of learning the Hamiltonian gradients that involve limit state function (model) evaluation. This way, model evaluation is only required during HNN training and in the last (acceptance/rejection) step. This makes H\textsubscript{NN}MC-based Subset Simulation substantially faster computationally. 

\section{Proof of concept}\label{sec:PoC}

In this section, we provide a rigorous proof of concept presenting several numerical examples of reliability problems using the proposed method. For the chosen examples, we use the reliability index $\beta=\Phi^{-1}(P_F)$ and compare the results for accuracy with the modified Metropolis-Hastings (MMH) sampler \cite{au2001estimation} and for computational performance with existing HMC methods \cite{wang2019hamiltonian}. Using the reliability index, instead of the probability of failure itself, allows us to make meaningful comparisons of the coefficient of variation, which can break down for $P_F$ \cite{shields2021subset}. The problems are implemented in Python using the UQPy package \cite{UQpy,tsapetis2023uqpy}. Parts of the code for latent Hamiltonian Neural Networks have been adopted from the GitHub repository  Bayesian inference with Hamiltonian Neural Networks (BIhNNs)  available at (\href{https://github.com/IdahoLabResearch/BIhNNs}{https://github.com/IdahoLabResearch/BIhNNs}). The computational time of the examples was assessed on an Apple MacBook Pro equipped with an M1 chip and 16 GB of memory.

For all examples, we use $10^3$ samples from the original distribution and $10^3$ samples in each conditional level within the rejection sampling Subset Simulation as proposed by Wang et al. \cite{wang2019hamiltonian}. The conditional probability is fixed at $0.1$. Unless otherwise specified, $4\times 10^5$ gradient evaluations of the Hamiltonian are used to generate the training data set for the HNN, which is composed of three hidden layers and 100 neurons in each layer. In all cases, these settings give reasonable results. However, we have not performed a more rigorous network optimization on the HNN. Thus, performance may be improved through a better network design aimed at optimizing the learning rate. This problem is beyond the scope of this paper since it is independent of the approaches proposed here.

\subsection{Linear Limit State with Degenerate Gaussian Distributions}
\begin{figure*}[!ht]
    \centering
    \includegraphics[width = \textwidth]{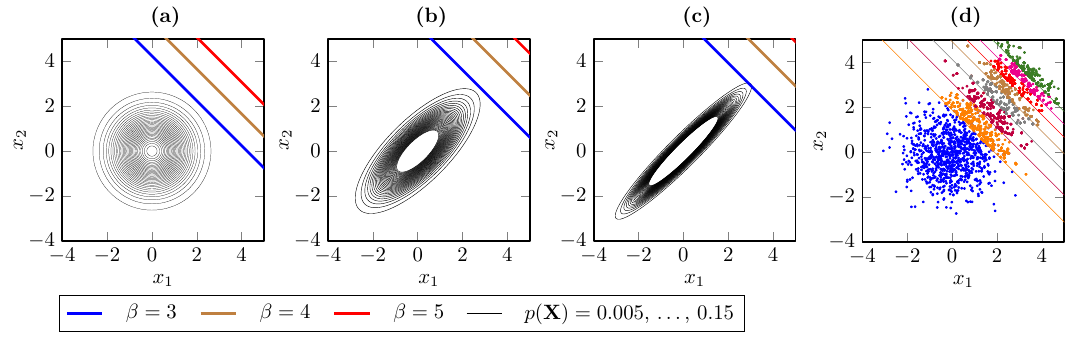}
    \caption{Illustration of the reliability problem for a bi-variate normal distribution with linear limit state function for different correlations: (a) $\rho=0$; (b) $\rho=0.75$; (c) $\rho=0.95$. (d) Illustration of the samples generated from Subset Simulation using the proposed H\textsubscript{NN}MC for a two-dimensional uncorrelated standard normal random vector having linear limit state function with $\beta=5$.  }
    \label{fig:Subset_problem_corr_gauss}
\end{figure*}

The first numerical example aims to explore the performance of H\textsubscript{NN}MC-based Subset Simulation for problems where the distribution form is simple, but sampling is difficult due to the degeneracy of the distribution. That is, the distribution is defined in a space with dimension $D$, but its support lies primarily on a space with dimension $d<D$. For this, we consider a random vector $\bm{X}$ with multivariate normal distributions having mean vector and covariance matrix given by:
\begin{equation}
\boldsymbol{\mu} = \left[ 0,0, \cdots, 0,0 \right]\text{;} \qquad  \mathbf{C}=
\begin{bmatrix}
1       &\rho &\ldots  &\rho\\
\rho & 1      &\ddots  &\vdots\\
\vdots  &\ddots  &1       &\rho\\
\rho &\ldots  &\rho &1
\end{bmatrix} \;.
\end{equation}
where the correlation $\rho$ is modified to control the dependence between the random variables. Importantly, as $\rho\to1$ the distribution degenerates and becomes effectively one-dimensional, sampling using standard MCMC methods (e.g.\ Metropolis-Hastings) becomes a challenge. The linear limit state function is given by $g(\mathbf{x}) = \beta \sqrt{\sigma_{max} n} - \sum_{i=1}^n x_i $, 
where $\beta = \{3,4,5\}$ is the reliability index, $n$ is the problem dimension, and $\sigma_{max}$ is the largest eigenvalue of $\mathbf{C}$. The reliability problem is illustrated for $n=2$ in Figure~\ref{fig:Subset_problem_corr_gauss}, which shows the contours of the joint pdf and the limit state function for different values of $\beta$ and correlation of (a) $\rho=0$, (b) $\rho=0.75$, and (c) $\rho=0.95$. 

The generated samples for one Subset Simulation run using the proposed H\textsubscript{NN}MC with $n=2$, $\beta=5$, and $\rho=0$ are shown in Figure ~\ref{fig:Subset_problem_corr_gauss}(d). 
The acceptance rate of the samples within the conditional levels varies between $0.96$ and $0.99$ for the HMC acceptance criteria and between $0.16$ and $0.52$ for the subset rejection criteria.  The predicted probability of failure is $3.12\times 10^{-8}$ which corresponds to $\beta = 5.41$.

This example serves as a first demonstration that the proposed H\textsubscript{NN}MC can be used to estimate small failure probabilities such that we can test it on more challenging cases where existing methods encounter difficulties. Considering different correlations $\rho$ and dimensions $n$, we demonstrate its performance by repeating each case $100$ times to estimate the coefficient of variation. 

\begin{figure*}[!ht]
    \centering
    \includegraphics[width = \textwidth]{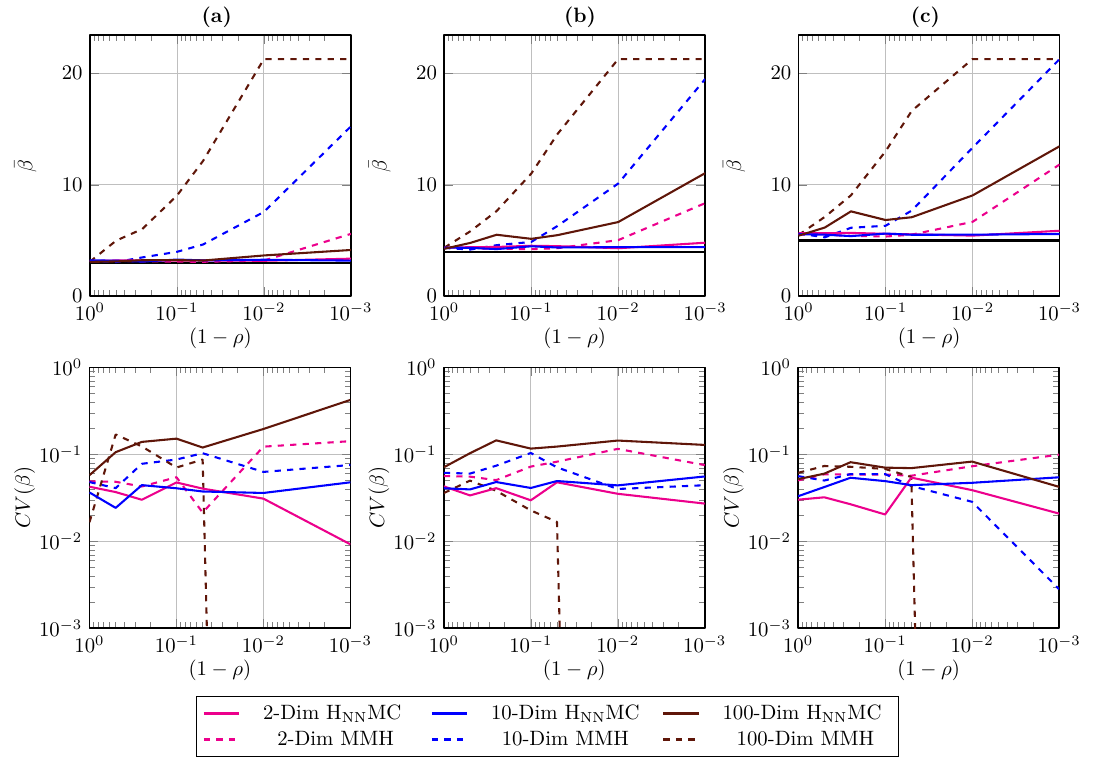}
    \caption{Results of 100 Subset Simulations using MMH and H\textsubscript{NN}MC for correlated Gaussian distributions: the mean reliability indices are compared in the upper plots while the lower plots show the coefficient of variation for (a) $\beta = 3$, (b) $\beta = 4$, and  (c) $\beta = 5$.}
    \label{fig:Subset_res_corr_gauss}
\end{figure*}

The results of the Subset Simulations using MMH and H\textsubscript{NN}MC are summarized in Fig.~\ref{fig:Subset_res_corr_gauss}, which shows the mean value and coefficient of variation of beta for different values of $\rho$ (plotted in log scale in terms of $1-\rho$) for low ($n=2$), medium ($n=10$), and high-dimensional ($n=100$) cases. As expected, the MMH gives good estimates for the failure probability for uncorrelated random variables, regardless of the dimension. However, as the correlation increases the MMH quickly loses accuracy, especially in high dimensions. In contrast, the H\textsubscript{NN}MC sampling strategy gives very good results in almost all problems and only starts to lose accuracy when $n=100$ and $\beta\ge 4$.
For the two and ten-dimensional correlated distributions, the H\textsubscript{NN}MC sampler gives very accurate results such that solid lines in the upper plots of Figure~\ref{fig:Subset_problem_corr_gauss} (mean values) are almost flat at the correct $\beta$ value. Furthermore, the coefficients of variation of the estimators from H\textsubscript{NN}MC are relatively low and only increase mildly with dimension. Meanwhile, for the 100-dimensional distribution, the CoV from MMH drops to zero as the correlation increases meaning that the method gives the wrong answer with little to no variability.

Notably, for this example, the primary objective is to demonstrate that the H\textsubscript{NN}MC can be used to achieve accurate reliability estimates. Computational efficiency is not the primary concern.
Nonetheless, the computational efficiency of using the proposed H\textsubscript{NN}MC method in Subset Simulations is compared with standard HMC in \ref{sec:Appendix_Degenerate_Gaussian}.

\subsection{Two-dimensional Rosenbrock distribution} \label{section:Rosenbrock}

For this example, we study a problem with two random variables following a Rosenbrock distribution having the unscaled probability density function written as:
\begin{equation}\label{eq:pdf_Rosenbrock}
    p(\mathbf{x}) \propto \frac{k \left(x_{2}-x_1^2\right)^2+\left(1-x_1\right)^2}{20}\;\text{.} 
\end{equation}
The distribution has a banana or boomerang shape, and depending on the factor $k$, the distribution can be wide or very thin. 
We define a linear limit state function given by $g(\mathbf{x})=120-x_2-3x_1$.
For this problem, the MMH algorithm can give reliable failure probabilities only for $k \le 1$ \cite{shields2021subset}. 

\begin{figure*}[!ht]
     \centering
     \begin{tikzpicture}
        \node[anchor=mid] at (-2.4,6.){\textbf{(a)}};
        \node at (-3.,3){\includegraphics{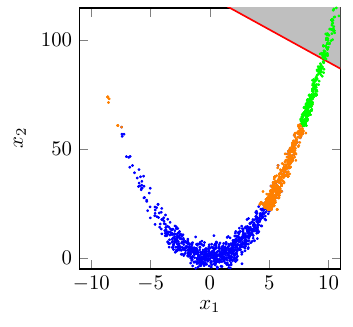}};
        \node[anchor=mid] at (3.6,6.){\textbf{(b)}};
        \node at (3.,3){\includegraphics{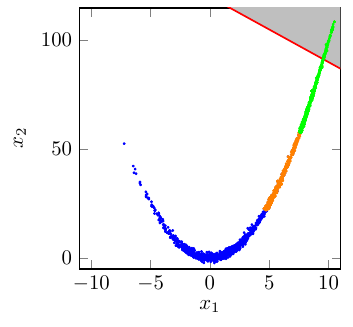}};
     \end{tikzpicture}
    \caption{Subset results for the Rosenbrock distribution using H\textsubscript{NN}MC: (a) Samples from each subset with $k=1$; (b) Samples from each subset with $k=10$.
    }
    \label{fig:Subset_rosenbrock_k10}
\end{figure*}

We first consider the Rosenbrock distribution with $k=1$ having ``true'' reliability index of $\beta =2.706$ ($P_F = 3.4 \times10^{-3}$) obtained from Monte Carlo simulations. Samples from one Subset Simulation using H\textsubscript{NN}MC are shown in Figure~\ref{fig:Subset_rosenbrock_k10}(a) yielding very accurate results with $\beta = 2.808$ ($P_F = 2.48 \times10^{-3}$). 
Using the same architecture for 100 Subset Simulations trials yields the mean reliability index $\Bar{\beta} = 2.953$ 
and coefficient of variation $CV(\beta)=0.0932$. 
To further test the approach, we increase $k$ to $10$, where the MMH algorithm fails to adequately sample from the failure region and therefore Subset Simulation with MMH breaks down. Figure~\ref{fig:Subset_rosenbrock_k10}(b) shows one subset using the H\textsubscript{NN}MC approach yielding $\beta = 3.039$ ($P_F =1.18 \times 10^{-3}$), which compares favorably with the ``true'' $\beta = 2.6755$ ($P_F =3.73 \times 10^{-3}$). 
The mean reliability index from 100 trials is $\Bar{\beta}= 3.435$ 
and the coefficient of variation is $CV(\beta)=0.127$, which compare favorably to state-of-the-art methods such as the affine invariant stretch sampler,
which gives $\Bar{\beta}= 3.466$, $CV(\beta)=0.138$ \cite{shields2021subset}. Some additional remarks on computational cost for this example are provided in \ref{sec:Appendix_Degenerate_Gaussian}.

\subsection{Limitations of the H\textsubscript{NN}MC in Subset Simulations for multi-dimensional Gauss and Rosenbrock distributions}

The proposed H\textsubscript{NN}MC algorithm reaches its limitations for strong correlation with high dimension and low probability of failure, e.g., the 100-dimensional problem with a correlation of $\rho = 0.999$ and $\beta = 5$.  The same applies to Rosenbrock distributions with high $k$.  While the HMC method still gives reasonable results, the proposed H\textsubscript{NN}MC faces higher rejection rates in the rare event space since it is particularly challenging to learn the steep gradients of the Hamiltonian correctly. In Section~\ref{section:error_monitoring} and~\ref{section:param_update}, we propose two strategies to mitigate this issue: (i) updating the neural network parameters in each subset and (ii) an online error monitoring scheme that deploys conventional HMC as needed. The limitations and strategies are demonstrated in \ref{sec:Appendix_Rosenbrock} where both approaches are applied to the two-dimensional Rosenbrock distribution with $k=100$. 

\subsection{Additional examples for proof of concept}

Additional examples related to structural dynamics are provided in the appendix. The reliability problems focus on system parameters, following different distributions in \ref{sec:Appendix_spring_damper} and uncertainty introduced within the load with high dimensional problems in \ref{sec:Appendix_white_noise}. As shown in these appendices, the proposed H\textsubscript{NN}MC Subset Simulation provides accurate estimations of the failure probability in both examples.

\section{Subset Simulations for systems with uncertain parameters}\label{sec:SS_BI}

The major advantage of the proposed method arises when model evaluations are necessary for Hamiltonian gradient evaluation. As highlighted by Wang et al.~\cite{wang2019hamiltonian}, this is not generally the case for reliability problems on deterministic systems having uncertain inputs with well-known distributions. However, when performing reliability analysis on a system whose parameters must be inferred from data, we cannot avoid performing model evaluations at each step of HMC. That is, model evaluation is \textit{necessary} for gradient evaluation. 

Furthermore, we cannot use automatic differentiation tools since model evaluation is required. Therefore, we use a central difference scheme to calculate the numerical gradients. Hence, the evaluation requires multiple model evaluations for each gradient computation, which becomes computationally excessively expensive for high dimensional problems. 

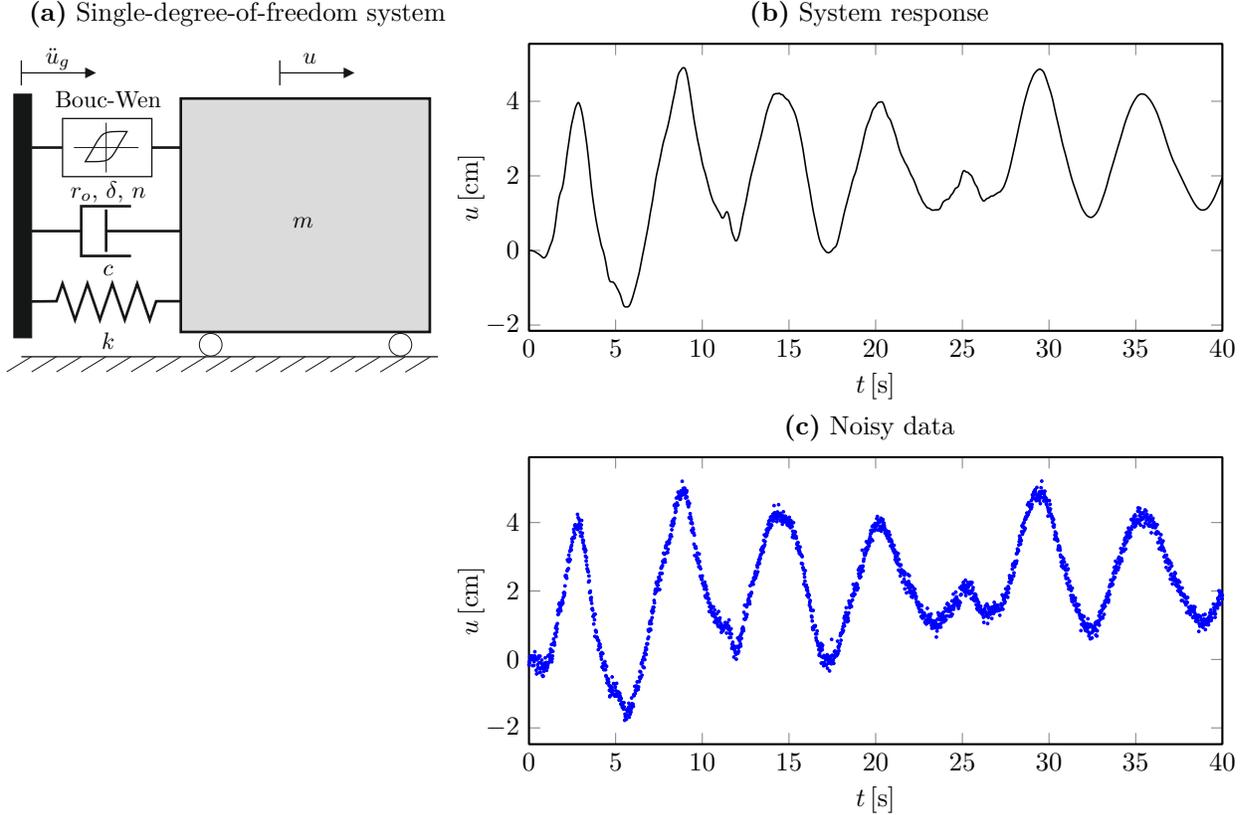
\begin{figure}[!ht]
    \centering
        \centering
     \begin{tikzpicture}
        \node[anchor=mid] at (-2.9,5.8){\textbf{(a)} Single-degree-of-freedom system};
        \node at (-3.1,3.2){\includegraphics[width=0.36\textwidth]{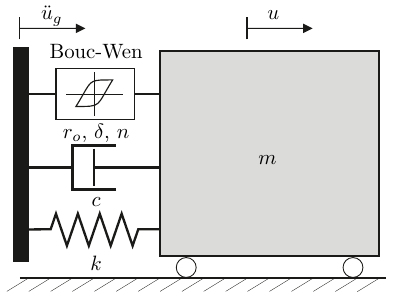}};
        \node[anchor=mid] at (5.5,5.8){\textbf{(b)} System response};
        \node at (5.2,3){\includegraphics{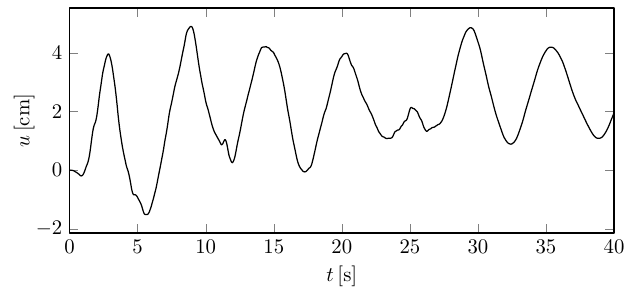}};
        \node[anchor=mid] at (5.5,.3){\textbf{(c)} Noisy data};
        \node at (5.2,-2.5){\includegraphics{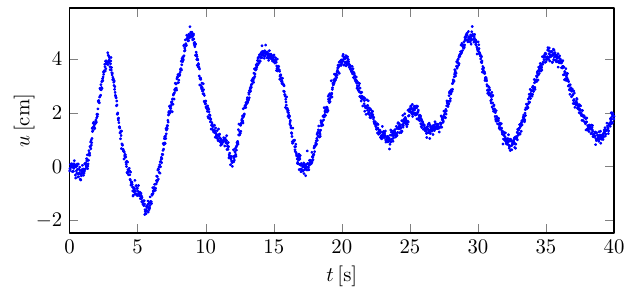}};
    \end{tikzpicture}

    \caption{Bouc-Wen model for Bayesian Inference}
    \label{fig:Bouc_wen}
\end{figure}

To illustrate this case, we apply the proposed H\textsubscript{NN}MC-based Subset Simulation approach to assess the reliability of a single-degree-of-freedom system having a Bouc-Wen hysteretic material model (illustrated in Figure~\ref{fig:Bouc_wen}a) whose parameters must be inferred from data.
The equation of motion of the  highly nonlinear system is written as~\cite{UQpy}:
\begin{align}
        & m \ddot{u}(t)+ c \dot{u}(t)+ k r(t) = - m \ddot{u}_g(t) \; \text{, \quad with} \\
        & \dot{r}(t) = \dot{u} - \beta \vert \dot{u} \vert \, \vert r(t) \vert ^{n-1} \, r(t) - \gamma \dot{u}(t) \vert r(t)\vert^n\; \text{.}
\end{align}
where $k$ is the spring stiffness and $n$ $\beta$ and $\gamma$ are the hysteresis parameters of the Bouc-Wen model. Alternatively, we parameterize the system equivalently by $n$, $r_0$ and $\delta$, where $r_0= \sqrt[n]{\dfrac{1}{\beta+\gamma}}$ and $\delta = \dfrac{\beta}{\beta + \gamma}$. 



The parameters of the spring and the Bouc-Wen material model are assumed to be unknown and must be learned from data. These data are generated by creating an artificial measurement of the system response to the scaled El Centro ground motion record~\cite{ElCentro} shown in Figure~\ref{fig:Bouc_wen}(b) with parameters $k=1 \frac{\text{N}}{\text{m}}\text{, }r_0=2.5 \text{ cm}\text{, }\delta=1,\text{, and }n=2$. The original system uses small viscous damping, which introduces a small error since the inference model assumes no damping. The simulated noisy response measurement, including $5\%$ root mean squared noise, is shown in Figure~\ref{fig:Bouc_wen}(c).

\begin{figure}[!ht]
    \centering
    \includegraphics[width=\textwidth]{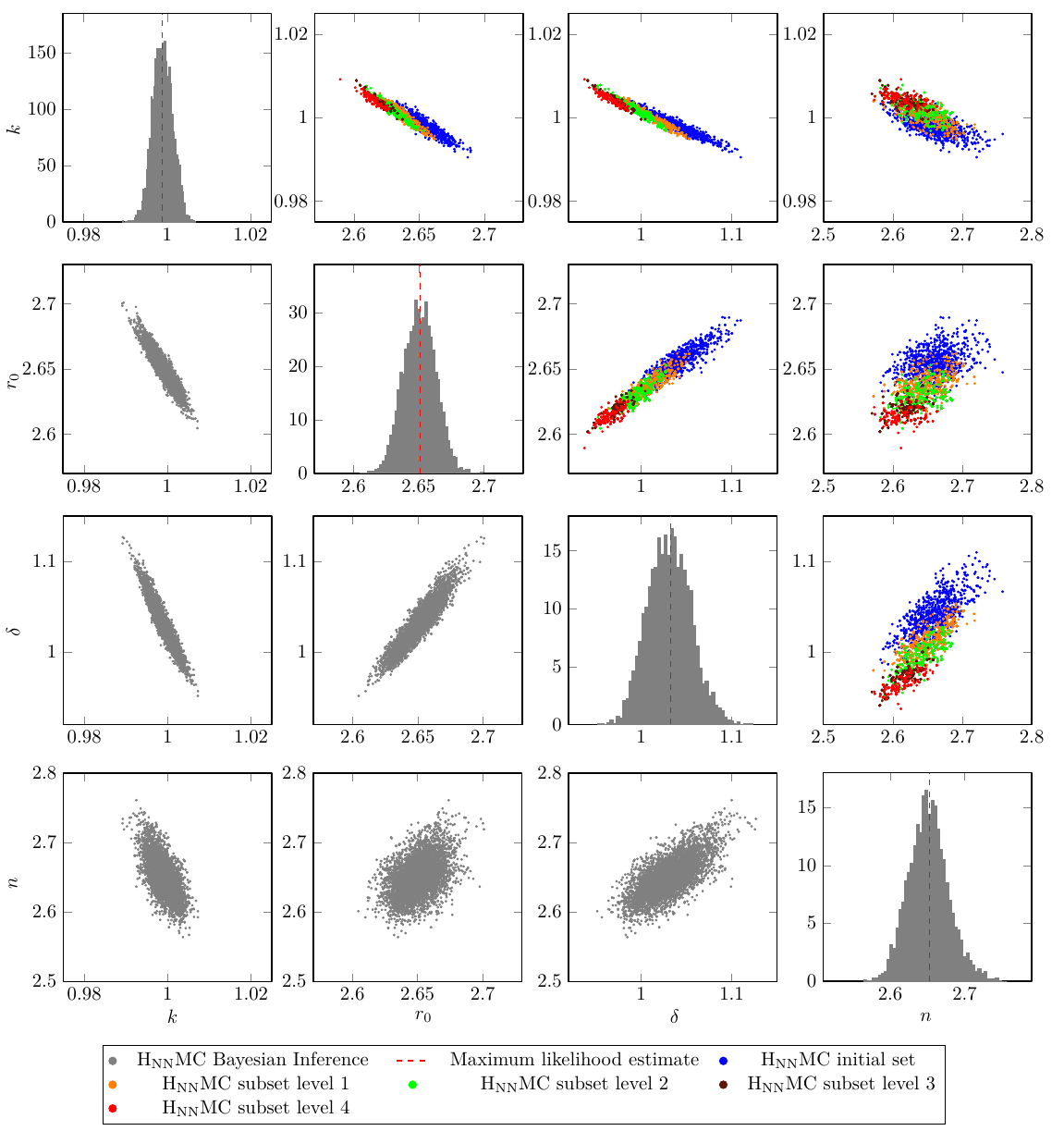}
    \caption{Subset Simulations for Bayesian inference; Bottom left: Drawn samples using H\textsubscript{NN}MC in Bayesian inference sampling; Diagonal: Histograms of $5000$ samples using H\textsubscript{NN}MC; Top right: Subset Simulations using H\textsubscript{NN}MC with $1000$ samples. }
    \label{fig:BW_BI_HNN_results}
\end{figure}

Using the observed data, we first employ traditional HMC to infer the joint distribution of the material parameters. This HMC comes at a significant computational cost since each leapfrog step requires evaluation of the model. We draw a total of $5000$ samples by employing $50$ leapfrog steps for each proposal, gathering $250\,000$ gradient evaluation in total. We then use these gradient evaluations to train the HNN -- hence the HMC cost is not wasted. 
For reference, the H\textsubscript{NN}MC is tested on the Bayesian inference problem first, where we again generate $5000$ samples from the posterior parameter distribution. 
These samples, which are generated extremely fast because they do not require model evaluation for the Hamiltonian gradients, are shown in the bottom left plots of Figure~\ref{fig:BW_BI_HNN_results}, and histograms for each variable are shown on the diagonal. The observed results show that the H\textsubscript{NN}MC distributions are consistent with those observed from previous studies~\cite{UQpy}. 


Next, we conduct reliability analysis using the inferred joint parameter distribution. 
The limit state function corresponds to the exceedance of a displacement limit, $u_{lim}=5$, in the time range $t\in[0, 40]$ sec under the same El Centro excitation and is expressed as:
\begin{equation}
    g(\mathbf{x}) = u_{lim} -   \max_{t \in [0,40]} \vert u(\mathbf{x},t) \vert \; \text{.}
\end{equation}
where $\mathbf{x}$ is the vector of uncertain system parameters.
We apply the H\textsubscript{NN}MC sampling strategy within Subset Simulation using $1000$ samples per subset. Repeating the analysis for $100$ trials, we observe a mean reliability index of $\bar\beta=4.35$ with coefficient of variation $CV(\beta)=0.048$. A representative sample set from Subset Simulation with samples colored by subset is shown in the upper off-diagonal plots in Figure \ref{fig:BW_BI_HNN_results}.

The H\textsubscript{NN}MC results are in agreement with Subset Simulations performed with HMC using the traditional gradient updates, which produced a mean reliability index of $\bar\beta=4.44$ with coefficient of variation $CV(\beta)=0.039$ from ten independent trials. However, the cost of Subset Simulation with H\textsubscript{NN}MC is dramatically lower. Traditional HMC requires that the system response be evaluated at \textit{every} time step of the Hamiltonian trajectory. Therefore, one Subset Simulation takes $t_{\text{HMCSS}}=579\,549 \text{ sec }(>  6 $ days). As a result, we could only afford 10 independent trials with standard HMC. Because the trained HNN automatically computes the gradients, and no model evaluation is needed, The H\textsubscript{NN}MC method takes only  $t_{\text{HMCSS}}=4509 \text{ sec}$  ($\approx 1.25 $ hours) per Subset Simulation -- an improvement of more than 100x.


\section{Conclusion}\label{sec:Conclusion}

In this paper, we developed Hamiltonian Neural Network Markov Chain sampling for reliability analysis using Subset Simulations. The proposed strategy reveals high efficiency and accurate estimates for the probability of failure.

Hamiltonian Neural Networks are particularly powerful for gradient predictions of long trajectories \cite{Greydanus2019}, so marriage with Hamiltonian Monte Carlo is highly beneficial. As a result of this, the acceptance rate of the non-random walk proposals remains high.

The main benefit of using this approach is to decrease the computational efforts for the gradient calculation during the sampling. For the examples in this paper, the new approach is at least $20$ times faster when performing Subset Simulations. However, a fair metric for the speed-up of the approach also considers the training procedure of the Hamiltonian Neural Network, which includes the traditional gradient calculation during the generation of the training set. The gradient evaluations required for the Subset Simulations are -- dependent on the settings -- far more than the number of samples for the training set. Notably, the strategy becomes increasingly beneficial if the same distribution is used for several simulations since the training procedure is only required once.

In the case of very complex and high-dimensional distributions, such as the Rosenbrock and the highly correlated 100-dimensional Gaussian distribution, the approach reaches its limitations in higher subset levels. The most reliable solution for this problem is to include an online error monitoring scheme. Although this extension weakens the computational efficiency of the H\textsubscript{NN}MC subset method proposed, online error monitoring enables comparable results to the standard HMC method, even in the most unfavorable situations, i.e., for very complex distributions. However, for most updates, the online error scheme is not required.

The H\textsubscript{NN}MC proposed in this paper is particularly valuable when applied to problems where the underlying distribution is \textit{a priori} unknown. For such Bayesian inference problems, the numerical gradient calculation becomes extremely expensive when applying the traditional HMC method since the underlying model has to be evaluated. These gradient evaluations are done by the Hamiltonian Neural Network in a fractional amount of time, enabling outstandingly fast and efficient sampling. Thus, the algorithm proposed in this paper reveals the highest speed up in the Bayesian inference framework since it decreases the computational effort to $1\%$ compared to the time necessary using the traditional HMC sampler.

\section*{CRediT authorship contribution statement}

\textbf{Denny Thaler}: Conceptualization, Methodology, Software, Data curation, Writing- Original draft preparation, Funding acquisition. \textbf{Somayajulu LN Dhulipala}: Methodology, Writing- Reviewing and Editing, Funding acquisition. \textbf{Franz Bamer}: Conceptualization, Supervision, Writing- Reviewing and Editing. \textbf{Bernd Markert}: Writing- Reviewing and Editing. \textbf{Michael D. Shields}: Conceptualization, Methodology, Supervision, Writing- Reviewing and Editing, Funding acquisition.

\section*{Declaration of interests}
The authors declare that they have no known competing financial interests or personal relationships that could have appeared to influence the work reported in this paper.

\section*{Acknowledgement}
  Denny Thaler was supported by a fellowship of the German Academic Exchange Service (DAAD) during his research stay at Johns Hopkins University.  Somayajulu Dhulipala and Michael Shields were supported through INL's Laboratory Directed Research and Development (LDRD) Program under U.S. Department of Energy (DOE) Idaho Operations Office contract no. DE-AC07-05ID14517. Franz Bamer was supported by the Federal Ministry of Education and Research (BMBF) and the Ministry of Culture and Science of the German State (MKW) under the Excellence Strategy of the Federal Government and the Länder.

\bibliography{main}

\appendix
\counterwithin{figure}{section}
\section{Computational efficiency of H\textsubscript{NN}MC compared on proof of concept examples}\label{sec:Appendix_Degenerate_Gaussian}
\subsection{Degenerate Gaussian}

The computational cost of H\textsubscript{NN}MC for Subset Simulation is compared with standard HMC based on the number of Hamiltonian gradient evaluations for cases with $n=100$, $\rho=0$, and $\beta=3, 5$ in Table \ref{tab:corr_gauss_gradient eval}. Clearly, both H\textsubscript{NN}MC and HMC are capable of drawing quality samples in this uncorrelated high-dimensional case. The H\textsubscript{NN}MC only requires traditional gradient evaluations for the training data. Here, we used $4\times10^5$ gradient evaluations. It can then be directly deployed for all problems based on this distribution without any additional computations. Meanwhile, the traditional gradient updates using HMC highly depend on the settings of the algorithm. Using as few as possible gradient evaluations, i.e., one chain with $100$ burn-in steps, results in the number of gradient evaluations presented in Figure~\ref{tab:corr_gauss_gradient eval}. Hence, the proposed method only leads to a speed-up if the probability of failure becomes low. Of course, the H\textsubscript{NN}MC sampler is significantly more efficient when used for several trials since the HNN does not need to be retrained for each trial.

\begin{table}[!ht]
    \centering
    \caption{Comparison of the required Hamiltonian gradient evaluations for Subset Simulations using HMC and H\textsubscript{NN}MC for correlated Gaussian random variables with a linear limit state having $n=100$, $\rho=0$, and $\beta={3,5}$.
    }
    \begin{tabular}{c|c|c|c|c|c}
        Hamiltonian &          &         &        & number  & average \\
        gradient  &  training  & initial & subset & of      & for one  \\
        evaluations &   data   &  set    &        & subsets & simulation \\
         \hline 
         & & & & & \\[-1em]
        H\textsubscript{NN}MC  & $4\times 10^5$ & 0 & 0 & $2.8$/$7.5$ & $4\times 10^5$\\
        HMC ($\beta=3$) & 0 & $91\,516$ & $ 79\,856$ & $2.8$ & $315\,113$ \\
        HMC ($\beta=5$) & 0 &  $91\,075$ & $78\,282$  & $7.5$ & $678\,190$\\
    \end{tabular}

    \label{tab:corr_gauss_gradient eval}
\end{table}



Regarding computational time, the HNN prediction and the Hamiltonian gradient evaluation are compared for the uncorrelated Gaussian distribution ($\rho=0$) with $100$ variables. The numerical evaluation of $10^3$ gradients takes $2\,896\; \text{ms}$, whereas the HNN performs this task in $16 \;\text{ms}$ which is a reduction in cost by a factor of more than $180$. Assuming the HNN is already trained for this distribution, full Subset Simulations using the H\textsubscript{NN}MC run in $489$ sec on average, while the traditional HMC takes $30\,543$ sec on average.

\subsection{Rosenbrock}

The speed-up of the method for this example in terms of traditional gradient calculations depends on the number of samples used for the neural network training. For this distribution, using $8\cdot10^5$ gradients for training led to good accuracy during the supervised learning procedure. The traditional approach depends on the settings for the Hamiltonian Monte Carlo, especially since the initial set requires a relatively high burn length for this distribution to find adequate initial samples. The number of leapfrog steps is calculated using Algorithm~\ref{alg:init-t_f} with a fixed step length $\Delta t = 0.01$. Starting with $20$ independent samples from the Gaussian distribution and using a burn length of $500$ results in $881\,619$ traditional gradient calculations. The number of gradient evaluations could be further reduced if only one single chain is used, reducing the number of gradients required for the initial set to $120\,220$. However, within each subset, the chains operate independently, meaning that the number of required gradients is solely influenced by the trajectory and the step length. For the chosen example, the algorithm evaluated seven levels with the following gradient evaluations $\left[205\,200, 410\,364, 407\,220, 483\,184, 471\,440, 539\,776, 550\,668 \right]$, which is $3\,067\,852$ evaluations for all subset levels. Thus, the number of required conventional gradient evaluations is dependent on the number of chains chosen for the evaluation of the initial subset, but at least $4$ times higher compared to the HNN approach. 

To compare the computational time of the two approaches, we used $10^3$ samples in each subset, and the initial set of these examples requires the update of the $10^3$ samples in each run. The classic gradient evaluation takes $226.8 \;\text{ms}$, while the neural network requires only $10.9\; \text{ms}$. Thus, the neural network approach is approximately $20$ times faster for gradient evaluations. The evaluation of a whole Subset Simulation takes $15\,872$ sec on average for HMC and $735$ sec on average using H\textsubscript{NN}MC. Increasing the number of samples for the new strategy achieves better accuracy than using the HMC. Even though sampling is still faster -- using $10$ times more samples, $10^4$ instead of $10^3$ takes $5\,393$ sec on average -- the evaluation of the limit state function is in general the most expensive part of the simulation.

\section{Limitations for Rosenbrock distribution with $k=100$}\label{sec:Appendix_Rosenbrock}

\begin{figure}[!ht]
    \centering
     \begin{tikzpicture}
        \node[anchor=mid] at (-2.4,6.){\textbf{(a)}};
        \node at (-3.,3){\includegraphics{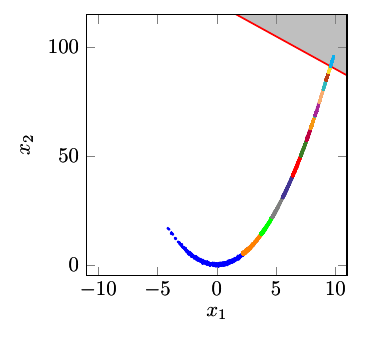}};
        \node[anchor=mid] at (3.6,6.){\textbf{(b)}};
        \node at (3.,3){\includegraphics{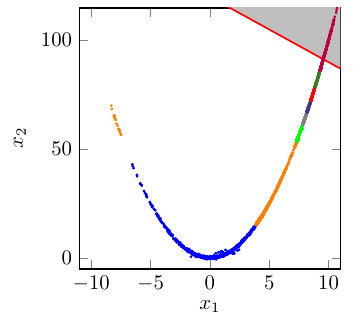}};
        \node[anchor=mid] at (-2.4,0.){\textbf{(c)}};
        \node at (-3.,-3){\includegraphics{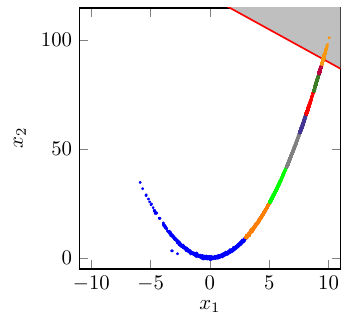}};
        \node[anchor=mid] at (3.6,0.){\textbf{(d)}};
        \node at (3.,-3){\includegraphics{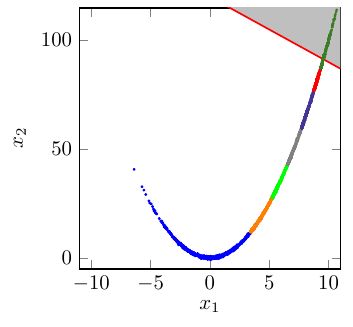}};
        \node[anchor=mid] at (-2.4,-6.){\textbf{(e)}};
        \node at (-3.,-9){\includegraphics{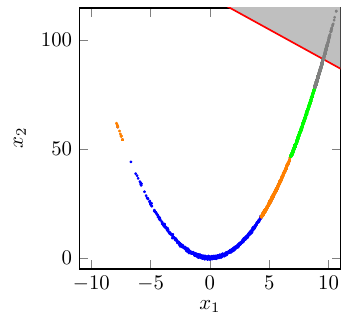}};
        \node[anchor=mid] at (3.6,-6.){\textbf{(f)}};
        \node at (3.,-9){\includegraphics{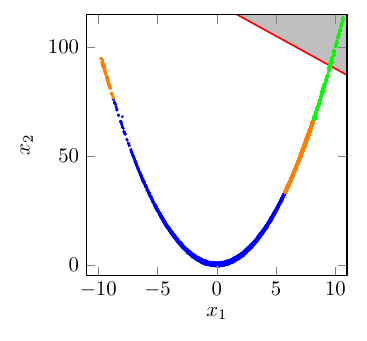}};
     \end{tikzpicture}
    \caption{Subset results for the Rosenbrock distribution $k=100$ using Hamiltonian Neural Network sampling (a) with a Hamiltonian Neural Network, (b) with a latent Hamiltonian Neural Network, (c) with a Hamiltonian Neural Network and retraining in each set, (d) with a Hamiltonian Neural Network and online error monitoring, (e) Hamiltonian Monte Carlo sampling, and (f) Hamiltonian Neural Network with $10\,000$ samples.
    }
    \label{fig:Subset_rosenbrock_k100}
\end{figure}
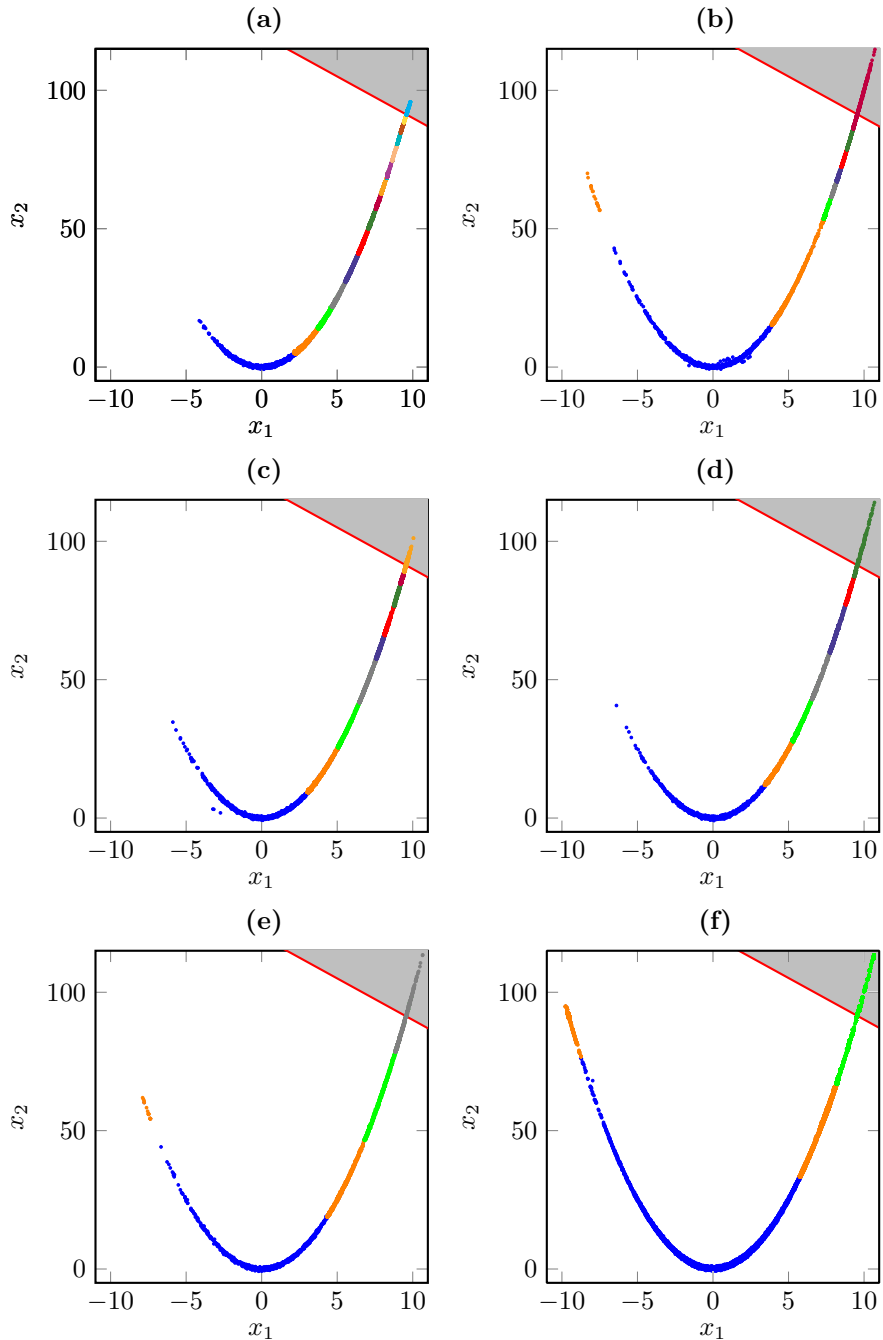

This part of the appendix focuses on the proposed strategies to deal with the limitations of the approach based on the Rosenbrock distribution with $k=100$. First, we observe that using the same training procedure as in previous cases ($4\times10^5$ initial gradient evaluations) results in very few samples in the tails of this very thin distribution. As a result, the HNN is likely to produce poor gradient predictions in this area -- which is critical for reliability analysis. Hence, we use long trajectories ($800$ leapfrog steps with a step size of $0.05$) during the creation of training data assembled using $10^3$ initial samples without using an acceptance-rejection criterion. 
Figure~\ref{fig:Subset_rosenbrock_k100}(a) shows the results of one Subset Simulation using H\textsubscript{NN}MC. The pre-trained HNN on $8\times10^5$ gradient evaluations gives accurate predictions within the first three sets; the acceptance rate is above $90\%$. However, the rate drops rapidly due to the poor training in the tails (e.g., the acceptance rate is $0.05$ in subset $8$ and does not exceed $0.25$ in any subset after the third). Hence, the samples reach the failure region only after $15$ subsets, which significantly underestimates the probability of failure. However, these observations represent only one trial. For further analysis, we used $10$ independently trained architectures to create 100 Subset Simulations, i.e., ten trials with each trained architecture. The results of these runs are summarized in Table~\ref{tab:results_subset_rosenbrock}. The standard HNN approach gives a mean beta value of $\Bar{\beta}=8.014$, which is not accurate relative the reference value $\beta = 2.715$ \cite{shields2021subset}.

To improve performance, we explore three strategies: 
\begin{enumerate}
    \item Using an enhanced HNN architecture, i.e., the latent variable HNN \cite{Dhulipala2022}.
    \item Updating the HNN parameters in each subset evaluation to achieve higher acceptance rates~cf. Section~\ref{section:param_update}.
    \item Using an online error monitoring scheme to ensure good proposals for the new states, cf. Section~\ref{section:error_monitoring}.
\end{enumerate}
All three approaches improve the estimation of the probability of failure, as shown in Figure~\ref{fig:Subset_rosenbrock_k100}(b)-(d) and Table~\ref{tab:results_subset_rosenbrock}. 

First, the enhancement of the architecture from HNN to latent HNN improves the prediction accuracy of the network \cite{Dhulipala2022}. Even though the latent HNN provides better gradient estimates, the results vary significantly with each simulation. The gradients are often overestimated in regions that are not trained well, as can be observed from Figure~\ref{fig:Subset_rosenbrock_k100}(b). In the lower region around $(0,0)$, we observe outliers stuck outside the high probability area. One interesting observation is the vast area covered by the second subset in orange. This approach slightly improves the mean value of beta but significantly reduces the variance (Table~\ref{tab:results_subset_rosenbrock}). 

The results can be further improved if more training data is used, particularly in the tails of the distribution. This leads us to the second approach: retraining the HNN after each subset. The results for a single Subset Simulation are shown in Figure~\ref{fig:Subset_rosenbrock_k100}(c). 
Although the probability of failure estimate is a little lower compared to the latent HNN prediction in this particular illustration, we observe that the outcome is better on average. Also, we observe higher acceptance rates within the subsets, gradually evolving toward the failure region. Table~\ref{tab:results_subset_rosenbrock}  
shows that the mean beta value decreases to $\Bar{\beta} = 5.615 $, which (while not particularly accurate) is comparable to estimates from other state-of-the-art samplers, such as the stretch sampling method~\cite{shields2021subset}. However, this approach requires the evaluation of samples using the traditional approach within each subset. Thus, the computational cost increases with the number of training samples chosen for each subset.

The best results of the proposed approaches are achieved by using online error monitoring. Here, the trajectory is corrected using traditional gradient calculations if a threshold is exceeded. Therefore, we observe that no samples get stuck outside of the distribution, see Figure~\ref{fig:Subset_rosenbrock_k100}(d). However, using this scheme diminishes the computational savings of the approach as it requires expensive computational gradient calculations when Hamiltonian conservation is violated. From Table~\ref{tab:results_subset_rosenbrock}, the mean beta value is $\Bar{\beta} = 4.983$, which is an improvement over other methods (except traditional HMC) for only $1000$ samples per subset~\cite{shields2021subset}. Furthermore, the variance of the approach is also reduced. This approach delivers the most reliable results for very complex distribution and low-probability regions. However, the threshold has to be chosen carefully. If chosen too large, the proposals may still be inaccurate. If chosen too small, for most of the samples, numerical gradient calculations will be used such that the approach ends up in the traditional HMC. 

Meanwhile, the traditional HMC approach produces accurate results with an average of $\beta = 2.997$ and a coefficient of variation of $CV(\beta)=0.032$ for ten Subset Simulations. Therefore, while the proposed improvements show better performance for this very difficult problem, they are still not capable of reproducing the accuracy of the full HMC approach.


\begin{table}
    \centering
        \caption{Subset Simulation results for different H\textsubscript{NN}MC approaches. Summary from 100 trials using ten different HNN architectures and ten repeated trials for each approach.}
    \begin{tabular}{l|c|c|c}
        Approach &   $\Bar{\beta}$   &  Var($\beta$)    & CV($\beta$) \\
         \hline 
         & & & \\[-1em]
        Standard HNN  & $8.014$ & $3.985$ & $0.249$ \\
        Latent HNN  & $6.519$ & $0.841$ & $0.140$ \\
        HNN with retraining & 5.615 & 1.658 & 0.229 \\
        HNN with online error  &  4.983 & 0.885 & 0.189 \\
        HMC (10 runs) & $2.997$ & $0.0932$ & $0.032$  \\
        HNN (10 runs, $10^4$ samples/subset) & $2.781$ & $0.0313$ & $0.064$ 
    \end{tabular}

    \label{tab:results_subset_rosenbrock}
\end{table}

Another possibility for increasing the accuracy of Subset Simulations is to evaluate more samples in the subsets. Using $10^4$ samples in H\textsubscript{NN}MC results in a crucial improvement of the estimate. From $10$ runs, the H\textsubscript{NN}MC results in an average estimate of $\beta = 2.781$ with a coefficient of variation of $CV(\beta)=0.063$. The H\textsubscript{NN}MC with increased subset samples outperforms the traditional method regarding speed and accuracy. However, if the limit state functions become computationally more expensive, the sample increase will lead to significantly higher costs. Thus, to keep a balance between sampling efficiency and the number of samples required online error monitoring is likely to be preferred.

\section{Spring damper system}\label{sec:Appendix_spring_damper}

\begin{figure}[!ht]
    \centering
    \includegraphics{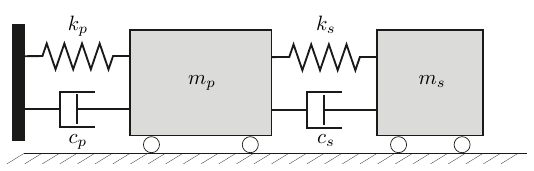}
    \caption{A two-dof system with uncertain parameters, cf~\cite{Sundar2016}.}
    \label{fig:2dof_spring_damper}
\end{figure}

In this example, the two-degree-of-freedom (dof) spring-mass-damper system shown in Figure~\ref{fig:2dof_spring_damper} is considered. The primary-secondary system is characterized by the masses $m_p$ and $m_s$, spring stiffnesses $k_p$ and $k_s$, and damping coefficients $c_p$ and $c_s$, where the subscripts $p$ and $s$ refer to the primary and secondary oscillators, respectively.
These six uncertain parameters follow independent log-normal distributions with mean values and coefficients of variation provided in Table~\ref{tab:settings_spring_damper}.
\begin{table}[!ht]
    \centering
    \caption{Mean values and coefficient of variation of the parameters of the two degrees of freedom system. All random variables of this example follow log-normal distributions. }
    \begin{tabular}{c|c|c}
        Parameter  &  Mean  & CV  \\
         \hline 
         & &  \\[-1em]
        $m_p$  &  $1.5 \,\text{kg}$ & $0.1$ \\
        $m_s$  &  $0.01 \,\text{kg}$ & $0.1$ \\
        $k_p$  &  $1. \,\frac{\text{N}}{\text{m}}$ & $0.2$ \\
        $k_s$  &  $0.01\, \frac{\text{N}}{\text{m}}$ & $0.2$ \\
        $\zeta_p$  & $ 0.05$ & $0.4$ \\
        $\zeta_s$  & $ 0.02$ & $0.5$ \\
        $F_s$  & $ 15\, \text{N}$ & $0.1$ \\
        $S_0$  & $ 100\, \text{N}$ & $0.1$ \\
    \end{tabular}

    \label{tab:settings_spring_damper}
\end{table}

The reliability of the system is based on the spring capacity $F_s$ of the secondary oscillator, where the highly nonlinear limit state function is given by~\cite{Kiureghian1991}:
\begin{equation}
g(\mathbf{x})=F_s-3 k_s \sqrt{\frac{\pi S_0}{4 \zeta_s \omega_s^3}\left[\frac{\zeta_a \zeta_s}{\zeta_p \zeta_s\left(4 \zeta_a^2+\theta^2\right)+\gamma \zeta_a^2} \frac{\left(\zeta_p \omega_p^3+\zeta_s \omega_s^3\right) \omega_p}{4 \zeta_a \omega_a^4}\right]} \;\text{,}
\end{equation}
where $\omega_p=\sqrt{{k_p}/{m_p}}$ and $\omega_s=\sqrt{{k_s}/{m_s}}$ denote the natural frequencies and $\gamma = {m_s}/{m_p}$ the mass ratio. Furthermore, $\omega_a= (\omega_p +\omega_s)/{2}$ and $\zeta_a = {(\zeta_p + \zeta_s)}/{2}$ denote the average values for the eigenfrequency and damping parameters, and the factor $\Theta$ is defined as $\Theta= {\omega_p - \omega_s }/{\omega_a}$.
The probability of failure is $P_F= 4.79\times10^{-3}$ ($\beta = 2.59$) determined from Monte Carlo simulation~\cite{Sundar2016, Kiureghian1991}. 

We performed $100$ Subset Simulations using the proposed H\textsubscript{NN}MC sampling. The resulting mean reliability index is $\Bar{\beta}=2.680$, and the coefficient of variation is $CV(\beta)= 0.0433$, which is very close to the correct value from MCS. 
We noticed that the acceptance rate of the subset drops below $0.3$, which would lead to an update of the leapfrog steps. The acceptance rate of the HNN proposals remains above $0.9$, indicating that the previously discussed online error monitoring and/or retraining are unnecessary.  The traditional HMC sampling for ten Subset Simulations results in a 
similar estimate with mean reliability index $\Bar{\beta}=2.663$ and the coefficient of variation $CV(\beta)=0.025$.

\section{White noise excitation}\label{sec:Appendix_white_noise}

\begin{figure}[!ht]
    \centering
    \includegraphics{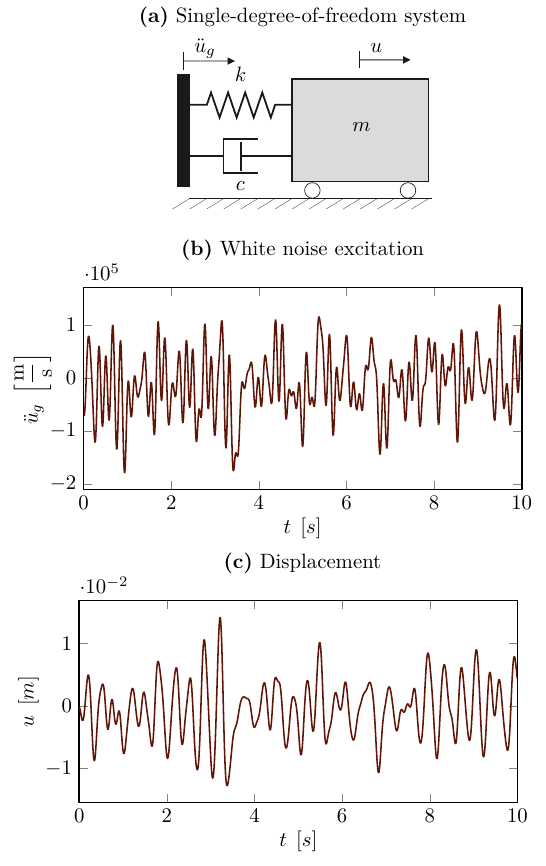}
    \caption{White noise excitation problem; (a) single-degree-of-freedom oscillator; (b) exemplary created white noise; (c) examplary response of the oscillator.}
    \label{fig:1dof_wn_struc_resp}
\end{figure}

This example considers a high-dimensional problem with $200$ variables. A single-degree-of-freedom oscillator, shown in Figure~\ref{fig:1dof_wn_struc_resp}(a), is excited with a seismic loading~\cite{wang2019hamiltonian}. The properties of the oscillator are fixed with mass $m=6 \times 10^4 \;\text{kg}$, stiffness $k = 2 \times 10^7\; \text{N}/\text{m}$ and damping $c= 2 m \zeta \sqrt{k/m}$ using a viscous damping ratio  of $\zeta = 0.1$. The equation of motion is written as:

\begin{equation}
    m \ddot{u}(t)+ c \dot{u}(t)+ ku(t) = - m \ddot{u}_g(t) \; .
\end{equation}

Here, $u(t)$ is the displacement of the oscillator, and  $\dot{u}$ and $\ddot{u}$ are the velocity and acceleration of the mass, respectively. The randomness of this example is introduced by the ground acceleration $\ddot{u}_g$. The acceleration is simulated by a white noise process discretized in the frequency domain as~\cite{Shinozuka1991}:

\begin{equation}
    \ddot{u}_g = A \sum_{j = 1}^{n/2} \left( x_j \cos{\omega_j t} + x_{j+n/2} \sin{\omega_j t}\right) \; ,
\end{equation}

where $A=\sqrt{2 S \Delta\omega}$ to account for the intensity of the white noise, chosen as $S = 0.01 \text{m}^2/\text{s}^3$. Figure~\ref{fig:1dof_wn_struc_resp}(b) shows one white noise sample.

The random vector $\mathbf{q}$ consists of $n=200$ independent standard Gaussian random variables. The frequency points are given by $\omega_j =j \Delta \omega $ using $n/2=100$ points with a cut-off frequency of $\omega_{cut}= 15\pi$, which leads to $\Delta \omega = 0.15 \pi$.

We use the same first-passage probability problems in the following, as demonstrated for the standard Hamiltonian Monte Carlo Subset Simulation~\cite{wang2019hamiltonian}. The limit state function of the problem is written as:

\begin{equation}
    g(\mathbf{x}) = u_{lim} -   \max_{t \in [0,10]} u(\mathbf{x},t) \; .
\end{equation}

The results of 100 Subset Simulations for $u_{lim}=0.02\; \text{m}$, $u_{lim}=0.025 \;\text{m}$ and $u_{lim}=0.03\; \text{m}$ in terms of the reliability index $\beta$ are shown in Table~\ref{tab:results_subset_SDOF_wn}. In the last column, the results from the original proposal of Hamiltonian  Monte Carlo for Subset Simulations are shown to compare the new method. 

\begin{table}[!ht]
    \centering
        \caption{Subset Simulation results of 100 runs of the Hamiltonian Neural Network approach for the white noise excitation subjected to a single-degree-of-freedom system. }
    \begin{tabular}{l|c|c|c|c}
        Threshold  &   $\Bar{\beta}$   &  Var($\beta$)    & CV($\beta$) & HMC \cite{wang2019hamiltonian} \\
         \hline 
         & & & \\[-1em]
        $0.020 \text{m}$  & $2.355$ & $0.0074$ & $0.0365$ & $2.471$ \\
        $0.025 \text{m}$  & $3.802$ & $ 0.0661$ & $0.0676$ & $3.786$\\
        $0.030 \text{m}$  & $5.165$ & $0.0775$ & $0.0539$ & $4.997$\\
    \end{tabular}

    \label{tab:results_subset_SDOF_wn}
\end{table}




\end{document}